\documentclass{article}
\usepackage[utf8]{inputenc}
\usepackage[margin=1in]{geometry}

\usepackage{times}
\usepackage[natbib=true,style=authoryear-comp,doi=false,isbn=false,url=false]{biblatex}
\bibliography{bib.bib}

\RequirePackage{amsmath,amssymb,amsthm,bm,bbm,fullpage,graphicx,psfrag,amsfonts,verbatim,mathtools,upgreek,xcolor}
\numberwithin{equation}{section}
\usepackage{subcaption}
\usepackage{thmtools}
\usepackage{thm-restate}
\usepackage[hidelinks]{hyperref}
\usepackage[ruled]{algorithm2e}
\usepackage{makecell}
\usepackage{tikz}
\usepackage{tikz-cd}
\usetikzlibrary{decorations.pathreplacing,calligraphy,bayesnet}
\usepackage{cases}
\usepackage{booktabs}
\usepackage{appendix}
\usepackage[nameinlink,noabbrev]{cleveref}

\DeclareMathOperator*{\argmin}{argmin}
\DeclareMathOperator*{\argmax}{argmax}

\theoremstyle{definition}
\newtheorem{definition}{Definition}[section]
\numberwithin{equation}{section}
\newtheorem*{theorem*}{Theorem}
\newtheorem{theorem}{Theorem}

\title{Achieving Representative Data via Convex Hull Feasibility Sampling Algorithms}
\author{Laura Niss, Yuekai Sun, Ambuj Tewari\\ University of Michigan}
\date{}

\begin{document}

\maketitle

\begin{abstract}
  Sampling biases in training data are a major source of algorithmic biases in machine learning systems. Although there are many methods that attempt to mitigate such algorithmic biases during training, the most direct and obvious way is simply collecting more representative training data. In this paper, we consider the task of assembling a training dataset in which minority groups are adequately represented from a given set of data sources. In essence, this is an adaptive sampling problem to determine if a given point lies in the convex hull of the means from a set of unknown distributions. We present adaptive sampling methods to determine, with high confidence, whether it is possible to assemble a representative dataset from the given data sources. 
    % We define the direction of greatest uncertainty and present three sampling policies that rely on this direction. 
    % Using sampling complexity as the performance measure, we provide some upper and lower bounds in the Bernoulli that show the inefficiency of uniform sampling compared to a simple lower-upper confidence bound (LUCB) interval policy. 
    We also demonstrate the efficacy of our policies in simulations in the Bernoulli and a multinomial setting.
\end{abstract}

\section{Introduction}

% [2) The motivating problem of representative sampling is somewhat obfuscated in the intro -- it seems that the second paragraph of the intro is really the central point, but it seems buried, and the problems described there are not emphasized or sign-posted as central. One step that could be taken would be to start the first paragraph (26-37) by directly identifying representation bias as the motivating problem, and de-emphasize the other sources of unfairness such as historical bias, which to my understanding do not motivate this paper. This clarity issue would also be partly addressed by the proposed change in 3.1.
% ]
Implementing algorithmic fairness in practice is a difficult task because most data science pipelines consists of many steps (e.g.\ data collection, data cleaning, training and post-processing), and any of these steps can affect the fairness of the outcome. Thus implementing algorithmic fairness in practice is generally non-trivial.
Representation bias is a known issue when training ML models \citep{hashimotoFairnessDemographicsRepeated2018b,rolfRepresentationMattersAssessing2021a}. This bias represent a lack of or minimal data from a subgroup of the desired population that can negatively impact the algorithmic outcomes. Unlike historical bias which is inherent in the data \citep{juliaangwinMachineBias2016a}, representation bias can be alleviated through intentional data collection.
When queried about ways individuals have attempted to address fairness, many cited more data collection as a first approach \citep{holsteinImprovingFairnessMachine2019a}. While this is possible in settings where group membership and, when applicable, outcome labels are known and can be directly sampled, there are circumstances where data collection comes from sources with unknown distributions of attributes. 

An example of this is given in \cite{holsteinImprovingFairnessMachine2019a}. Here they describe a company that wishes to automate essay scoring whose current iteration has unfair outcomes for a minority group. Their algorithm is scoring these minority students on average lower than a human specialist. They desire more high scoring essays from minority students to improve their scoring accuracy within that group. Because they do not know the distribution of these students at the schools they are collecting essays from, they do  not have an efficient strategy to collect those needed samples, or know if it is possible to collect a data set with their desired distribution. 

An approach to this problem would be to have a sampling policy to determine if there exists a distribution across schools that would produce a data set with the desired proportion of high scoring essay from the minority group. The goal of this iterative sampling policy would be to make this determination using a minimum number of samples. Once this feasibility is known, one can either sample accordingly or seek out other sources.

There are a myriad of strategies now published that are methods to improve fairness at the post-data collection stages \citep{dworkFairnessAwareness2012,friedlerComparativeStudyFairnessenhancing2019}. These training strategies and post-processing strategies will improve fairness outcomes, but there is a limit to improvement before impacting accuracy. It is always preferred in any machine learning application to start with the best data one can access. This highlights another benefit collecting fair data over post-collection strategies. If fairness is truly a concern, it must also be recognized that data collected today will be used for a different purpose tomorrow. By considering how to curate fair data in isolation, this can impact fairness outcomes regardless of the way data is used apart from its original purpose. 

For example, consider that, in general, different definitions of fairness cannot be simultaneously satisfied except for certain possibly unattainable scenarios \citep{kleinbergInherentTradeOffsFair2017,pleissFairnessCalibration2017a}. Collecting data to achieve one type of fairness when trained with a particular algorithm gives no guarantee for outcomes of other measures of fairness. If the measurement for fairness changes over the life of a project, the data is no longer optimal. Aiming for fair representation from the onset will mitigate some of these problems. Additionally, equal representation is one of the scenarios that can produce fair outcomes in relation to calibration and equalized odds, something that lopsided data cannot achieve.

This work aims to provide a sampling method that tests whether a curator can create a fair data set from available sources, where "fair" is defined in terms of a predefined proportion of group memberships. To the best of our knowledge, similar work in this area of fair sampling assumes a fair data set is achievable. This work focuses on testing that assumption. Considering the cost of collecting data, the goal will be to determine the feasibility of these sources with a minimum number of samples. When collecting data, if one can sample any protected attribute any number of times, it is simple to create training data that is consistent with some notion of fairness, such as equal proportions of protected attributes. In this paper, we consider the scenario where the sampling sources have unknown distributions of attributes and the curator has defined a "balanced" set in regards to the desired proportions of the training data. That is, data can be sampled from different sources (such as polling in different cities) but knowledge of the distributions of data from those sources is unknown. This problem setting is described in full in \cref{sec:problem definition}. 

Aside from collecting fair data for training, this method could also be used when fair sampling is the end for. For example, advertising community services with a desired outcome of equal men and women using those services. Different advertising strategies would reach different populations. A practitioner would want to know as quickly as possible whether their selected strategies can achieve their desired distribution, and if so what combination of strategies would do this.

% [1) It seems like the problem setting studied here is almost identical to that studied in source [23], but this was not perfectly clear. It would help to clarify this earlier, before you state your contributions, to make clear that your *approach* is new, but the problem setting is not.]

% [2) To me, the paper seems to center the approach of representing this sampling problem in a convex hull as a main contribution, e.g., line 82, "We introduce the convex hull feasible sampling problem". When a representation of a problem is a central contribution, I would expect it to fundamentally change the tractability of the problem (e.g., permit the application of methods established in that representation that could not otherwise be applied, change the set of tools available, or abstract away from extraneous information to simplify the problem). While the convex hull is a convenient representation of this (I think pre-existing, see 5.1?) problem, I could not discern from the paper why this approach is *fundamentally* useful to the problem in, e.g., any of the aforementioned ways. To me, the contribution is truly the work done on studying different sampling algorithms, but I may be missing something - either way, it would help if this were clarified.
% ]

\textbf{Contributions} We introduce the convex hull feasibility problem. In the Bernoulli setting, we give a lower bound on the expected sample size in the infeasible case and an oracle lower bound of the expected sample size in the feasible case. We define the direction of greatest uncertainty and present three policies that use this direction, along with a naive Uniform policy. Using high-probability upper bounds, we prove that one policy, Lower Upper Confidence Bound (LUCB) Mean is superior to Uniform. We define the Multinomial version of the problem along with adjusted algorithms, and using simulations show the performance of our three policies outperform Uniform in the Bernoulli setting and the Multinomial setting with three dimensions.

%%%%%%%%%%%%%%%%%%%%%%%%%%%%%%%%%%%%%
\subsection{Related Work}
\subsubsection{Fair Sampling}
To the best of our knowledge, the first work to address data collection as a part of bias mitigation is \cite{abernethyAdaptiveSamplingReduce2020a}. Here the goal is to optimize over both a loss function for accuracy and a loss function for fairness. They assume an infinite availability of group labeled data, and at every iteration of sampling they choose the sample which will either minimize the accuracy loss or minimize the fairness loss. The choice of which loss to minimize at every time point is determined by a Bernoulli variable with probability $p$, where $p$ is a parameter chosen beforehand. When a sample is chosen to increase fairness, a sample is drawn from the group which currently has the worst loss performance. Otherwise a sample is chosen randomly. The intuition in both cases is that more training samples will improve performance, either overall performance when sampling at random or a specific group's performance when sampling to improve fairness. A similar framework is presented in \cite{taeSliceTunerSelective2021}, where the groupings are predefined slices of a current data set, and the goal is to obtain additional samples within a budget so as to maximize average accuracy as well as minimize the average difference between the accuracy of each slice and that of the total data. Their sampling method relies on estimating learning curves and allocating the sampling budget to slices that will have maximum impact on accuracy and fairness.

Along this vein of work is \cite{shekharAdaptiveSamplingMinimax2021}. Their goal identify a minimax optimal classifier across the sampling proportion of protected attributes and the loss of the worst performing group. Given a function class $\mathcal{F}$, loss $l$, and protected attributes $z \in \mathcal{Z}$, they propose an adaptive sampling policy that identifies the worst performing group $z$ and dedicates a larger proportion of the sampling budget to that group. 

In \cite{asudehAssessingRemedyingCoverage2019}, they forgo optimization for a particular learning algorithm and focus on the coverage of features within the data. They define the set maximum uncovered patterns (MUP), which aims to identifying feature combinations that fail to meet predefined threshold counts. In addition to providing several algorithms to identify the set of MUP, they provide a greedy algorithm to sample additions data whose feature patterns are MUP until all meet the required sampling threshold.

The work closest to ours is presented in \cite{nargesianTailoringDataSource2021}, where the goal is to collect a data set of a given size consisting of a desired count from each defined group. Here they assume a priori that the desired counts are feasible, and if minimums are not achieved they propose oversampling until minority group counts are met and removing excess majority samples. In addition to results for when the sampling distributions are known, they tackle the unknown distribution model with a multi-armed bandit strategy. They propose a reward function that depends on the true distribution of a group (such as from census population data), with the intuition being if a sample is from a group with a high proportion in the population then the reward is low and if from a minority group the reward should be high. Using a UCB type strategy with this reward function presents a sampling strategy that aims to sample from the distribution with the largest proportion of the minority group. 
% When desired counts are not feasible, they propose oversampling until minority group counts are met and removing excess majority samples. 
Our work differs substantially by focusing on the feasibility of the desired proportions, and frames the problem through use of a convex-hull composed of points defined by a confidence region.

There are several other frameworks around obtaining a fair data set. For example, an active learning application is presented in \cite{anahidehFairActiveLearning2021}, where the goal is to sequentially select which points to label so as to balance model accuracy along with a predetermined notion of fairness. Data augmentation with synthetic points has also been explored \cite{sharmaDataAugmentationDiscrimination2020}.

\subsubsection{Bandit Pure Exploration}
The feasibility problem is closely related to the pure exploration multi-armed bandit problem. In pure exploration a learner has $k$ actions with unknown means and the goal is to identify the action or subset of actions with the largest mean from the fewest samples. There are two settings in this problem, fixed-confidence and fixed-budget. In the fixed-confidence setting, a policy aims to minimize the sample complexity while guaranteeing the outcome of a policy is correct with some minimum predetermined probability. In the fixed-budget setting, a policy, given a predetermined sample size, aims to provide the largest confidence with which the largest means are correctly identified. 

To see the connection to our feasibility problem to the fixed-confidence setting, consider the two class case, which reduces to identifying if there exists $p_i\leq x \leq p_j$. Here $p_1,\ldots,p_k$ are the $k$ unknown means and the desired mean $x$ encodes our definition of a balanced data set. Then by determining if $x$ is or isn't the maximum or minimum mean with some probability $1-\delta$ we determine whether we correctly identify feasibility with probability $1-\delta$. 

The PAC pure-exploration setting was first presented in \cite{even-darPACBoundsMultiarmed2002} for identifying the top action with a fixed confidence. Their successive elimination algorithm relies on uniformly sampling actions from a decreasing set, removing actions from the set as they are determined to be lower than the top action with high confidence. Another set of policies uses lower upper confidence bounds on the means of the actions  \citep{gabillonBestArmIdentification2012,kalyanakrishnanPACSubsetSelection2012,kaufmannInformationComplexityBandit2013,jamiesonLilUCBOptimal2014}. A lower bound on the expected sample complexity for Bernoulli rewards is presented in \cite{mannorSampleComplexityExploration2004}, where they provide worst case and gap dependent bounds. This is expanded upon by \cite{garivierOptimalBestArm2016}, who provide a lower bound on sample complexity for one parameter exponential families and a policy with a asymptotically matching upper bound. 

\subsubsection{Probabilistic Hyperplane Separability}
The fields of computational geometry and computer science are not new to the problems of convex hull feasibility and hyperplane separability with probabilistic points. Though the underlying data assumptions are not quite matched to the convex-hull feasibility problem we present in this paper, there are significant similarities that may ultimately be used in future research and we would be remiss not to point them out. The goal of these papers is typically to provide an algorithm identifying separability or the probability of separability that minimizes run time complexity.

We briefly characterize three variations of these problems that are similar to ours. The first is that which considers the probability of linear separability between two sets of points $A$ and $B$ which are drawn from sets $\mathcal{A}$ and $\mathcal{B}$, as in \cite{finkHyperplaneSeparabilityConvexity2017}. The second variation considers $n$ labeled points from sets $A$ and $B$, each with a known uncertainty region. The question then is to determine separability of sets of uncertainty regions, as seen in \cite{sheikhiSeparabilityImprecisePoints2017}. Finally, there is the problem formulation where there are $n$ points, with the value of each point $i$ having a probability distribution over a discrete set $s_i$ with the goal to find the probability a set $O$ lies within the probabilistic convex hull \citep{yanProbabilisticConvexHull2015}.

%%%%%%%%%%%%%%%%%%%%%%%%%%%%%%%%%%%%%
\section{General Problem Definition}
\label{sec:problem definition}

The fixed-confidence $\epsilon$-convex hull feasibility problem is defined as follows. Each of $k$ distributions, which we will hereto refer to as \emph{actions}, are independently belong to some known family $\mathcal{P}$ with unknown means $\mu_i$ in dimension $d$. We are given a known variable $x\in \mathbb{R}^{d}$ and a relaxation of $\epsilon \geq 0$ and define $x_{\epsilon}$ as the open set $\{ y : ||y - x|| < \epsilon\}$, with $x_{\epsilon} = x$ when $\epsilon = 0$.  
We define the \emph{feasible} case as  when there exists some $y \in x_{\epsilon}$ that lies in the convex hull of $\{\mu_1, ..., \mu_k \}$ and the \emph{infeasible} case as when the set $x_{\epsilon}$ lies outside of the convex hull of $\{\mu_1, ..., \mu_k \}$. We include the relaxation of $x$ with $\epsilon$ because it may not be necessary to achieve exact feasibility.

If the $\mu_i$'s are known, then it is possible to determine whether $x$ is in the convex hull of the $\mu_i$'s by solving a linear optimization problem. Instead, we consider the setting in which the $\mu_i$'s are unknown, but we can (actively) observe noisy versions of the $\mu_i$'s. The goal is to give a determination of the feasibility of $x_{\epsilon}$ with a predetermined confidence while minimizing the number of times the actions are sampled. 

In the fairness setting, the dimension $d$ represents the number of groups defined by the protected attribute labels that the curator wishes to balance on. For example $d=2$ could represent the groupings of `men' and `women'. The points $\mu_i$'s correspond to data sources: the $j$-th component of $\mu_i$ is the fraction of samples from the $j$-th group in samples from the $i$-th data source. The components of the query point $x$ correspond to the desired fractions of samples from each group in the data set. The convex hull feasibility problem is thus equivalent to determining whether there is a set of weights $w_i$ such that drawing $w_i$ fraction of samples from the $i$-th data source will lead to a data set with the desired fractions of samples from each group.

\subsection{Feasibility and Infeasbility}

Given $i \in [k]$, $\mu_i \in \mathbb{R}^{d}$, $x \in \mathbb{R}^{d}$ and $\epsilon \geq 0$, we first state the feasible and infeasible cases more formally.

\begin{definition}[Infeasible Case]
The problem is $(x, \epsilon)$-infeasible if there exists some separating hyperplane between $x_{\epsilon}$ and the $\mu_i$.
\[
    \exists a \in \mathbb{R}^d\ \text{such that} \ \forall i \in [k], x \in x_{\epsilon} \ (\mu_i-x)^Ta < 0. 
\]
\end{definition}

\begin{definition}[Feasible Case]
The problem is $(x, \epsilon)$-feasible if there exists a convex combination that expresses some $y \in x_{\epsilon}$ in terms of the $\mu_i$'s:
\[
    \exists \lambda \in \Delta^{k-1} \text{such that } y = \sum_{i=1}^k \lambda_i \mu_i.
\]
where $\Delta^{k-1}$ is the $(k-1)$-dimensional probability simplex in $\mathbb{R}^k$.
\end{definition}

Because the $\mu_i$ are unknown, we must rely on confidence regions to inform a decision of whether the underlying case is \emph{feasible} or \emph{infeasible}. If each confidence region $R_i$ contains $\mu_i$ with probability at least $1 - \frac{\delta}{k}$ then we can make a high-confidence decision on the underlying case.

\begin{definition}[$1-\delta$ Confident Infeasible]
There exists a separating hyperplane between the set $x_{\epsilon}$ and the confidence regions for all actions.
\[
    \exists a \in \mathbb{R}^d \ \text{such that} \ \forall i \in [k], \ y \in x_{\epsilon},\ v_i \in R_i,  \text{ we have that } (v_i-y)^Ta < 0.
\]
\end{definition}

\begin{definition}[$1-\delta$ Confident Feasible]
\label{def:df}
For all sets consisting of a point from each confidence region, there exists a point in $x_{\epsilon}$ within their convex hull. 
\[
    \forall v_i \in R_i, i \in [k], \ \exists \lambda \in \Delta^{k-1}, \ y \in x_{\epsilon}  \ \text{such that} \ y = \sum_{i=1}^k \lambda_i v_i.
\]
\end{definition}

\subsection{Sampling Policy}

A \emph{sampling policy} $\pi$ is a mapping of the history of all samples drawn up to the current time to the choice of which action to sample next and the termination of the algorithm. When a policy terminates, it outputs a result of either \emph{feasible} or \emph{infeasible}. Let $\tau$ represent the stopping time of a policy, and $I(\pi, \delta) \in$ \{\emph{feasible}, \emph{infeasible}\} be the indicator function of the output for policy $\pi$ given confidence $1-\delta$.

\begin{definition}[Sound Policy]
Given some $\delta$, We call a policy $(1-\delta)$-sound if the expected value of the stopping time is finite and if with probability at least $1-\delta$ the policy selects the correct underlying case, 
\[
E[\tau] < \infty
\]
\[
P(I(\pi, \delta) = feasible| feasible) \geq 1-\delta, \hspace{.5cm}
P(I(\pi, \delta) = infeasible| infeasible) \geq 1-\delta
\]
\end{definition}

%%%%%%%%%%%%%%%%%%%%%%%%%%%%%%%%%%%%%
\section{Bernoulli Feasibility Sampling}
We focus on the case where there are two protected categories ($d=2$). In this case the $\mu_i$ lie in the 2-dimensional simplex and convex-hull feasibility simplifies into testing in 1-dimension with Bernoulli means. This setting maps onto the scenario with two groups labels, $\{0,1 \}$, with $x \in [0,1]$ representing the desired proportion of samples from group 1 and the probability of sampling group $1$ from action $i$ is $p_i$. For our theoretical analysis, we assume without loss of generality that $p_1 \geq p_2 \geq ... \geq p_k$. 

\subsection{Sample Complexity Lower Bounds}
We will take inspiration from the pure exploration bandit literature and give a lower bound on the expected value of the stopping time $\tau$ as a measure of sample complexity in the Bernoulli setting.

The multi-armed bandit best arm identification problem and the Bernoulli convex hull feasibility problem share certain similarities pointing towards similar techniques, but significant differences prevent direct application. In the best arm identification problem, to determine the best action with high confidence, all sub-optimal actions must be sampled to some extent to rule them sub-optimal. This remains true in our problem when the problem instance is infeasible, as all actions must be sampled sufficiently to determine them separable from our set of interest $x_{\epsilon}$. If the instance is feasible, the relation of the "sub-optimal" actions to each other or $x_{\epsilon}$ becomes irrelevant. For example, if two actions are sampled such they are determined with high confidence to be above and below $x_{\epsilon}$ respectively, sampling from the other actions provides no additional information about the feasibility or infeasibility of the problem. Additionally, there may be multiple sets of actions whose convex hull is feasible. 

The possibility of multiple optimal subsets of actions presents a difficulty in determining a lower bound for feasible instances since for any $(1-\delta)$-sound policy, it may not have sampled all actions and there may be multiple sets of actions that would trigger termination with the correct outcome. Therefore, for a specific feasible instance, it becomes difficult to give an expected lower bound for each action, except for the case when the playable actions comprise a unique feasible set. 

Considering this, we give a looser oracle lower bound for the feasible case. Here, the oracle knows the optimal subset(s) of actions but does not know their means. The oracle lower bound then is the minimum expected sample complexity when only actions in an optimal subset are played. Note that the oracle lower bound is still a valid lower bound since we are only giving the learner more information about the problem. However, the true lower bound might be much higher than our oracle lower bound.

% [3) There were a few sentences introducing mathematical notation / ideas that I found vague:
% - In lines 281-282, {p1...pk} is a set of probabilities representing actions; I did not understand what was meant by a "feasible and infeasible instance of {p1,...pk}". For example, is "feasibility / infeasibility" with respect to a fixed set of data sources, and then we're taking the vector {p1,...,pk} as a point in space, and checking whether its in the convex hull of these data sources?
% - Lines 286 - 287: There are two feasible cases referenced throughout these sentences --- do these correspond to two different optimal actions, or two different sizes of J*? Would help to be explicit about what these cases are. I also do not see any reason for why, when |J*| = 2, J* is written as {1,k}. Does this correspond to the two optimal actions being p1 and pk (if so, it is not clear to me why this would be the case)?]

\textbf{Notation:} Let $\mathcal{E}_{f}(x,\epsilon), \  \mathcal{E}_{if}(x,\epsilon)$ represent the set of feasible and infeasible instances of $\{p_1,...,p_k\}$ for given $(x, \epsilon)$, respectively. Where a feasible instance represents a vector $\{p_1,...,p_k\}$ whose convex hull contains a point in $x_{\epsilon}$, and and infeasible instance is otherwise. For any feasible problem instance $\nu \in \mathcal{E}_{f}$, 
let 
$$\Omega = \{\mathcal{J} \subseteq [k] | \{p_i\}_{i \in  \mathcal{J}} \text{ is } (x, \epsilon) \text{ feasible} \}$$
be the set of all subsets of actions whose means are $(x, \epsilon)$-feasible. Then we define the optimal subset of actions, $\mathcal{J}^*$, as the subset(s) that is furthest from any infeasible instance, $\mathcal{J}^* = \argmax_{\mathcal{J} \in \Omega} \min_{\nu' \in \mathcal{E}_{if}} \sum_{i \in \mathcal{J}} D(\nu_i, \nu'_i)$, where $D$ is the Kullback–Leibler divergence. There are two feasible cases, either only one source is feasible or two sources are a feasible set, so $|\mathcal{J}^*|\in \{1,2\}$. When analysis differs for these cases and $|\mathcal{J}^*| = 1$ then we write $\mathcal{J}^* = \{l^*\}$, else we write $\mathcal{J}^* = \{1,k\}$, as in this case the optimal subset consists of the sources with the largest and smallest mean, $p_1$ and $p_k$.

\begin{theorem}[Oracle Feasible case]
\label{thm:Oracle Feasible case}
    For a problem instance $\nu$ that is $(x, \epsilon)$-feasible, for any $(1-\delta)$-sound deterministic policy with $d=2$, $\delta < 1/2$, 
    \begin{equation*}
        E_{\nu}[\tau] \geq 
    \begin{cases}
    \max \left\{ D(p_{l^*}|x-\epsilon)^{-1}, D(p_{l^*}|x+\epsilon)^{-1} \right\}
    \frac{1}{2}
    \log\left(\frac{1}{4\delta}\right) & \mathcal{J}^* = \{l^*\}\\
     \left[\frac{1}{D(p_1| x-\epsilon)} + \frac{1}{D(p_k|x+\epsilon)}\right]
     \frac{1}{2}
     \log\left(\frac{1}{4\delta}\right) & \mathcal{J}^* = \{1, k\}, 
    \end{cases}
    \end{equation*}
\end{theorem}

\begin{theorem}[Infeasible case]
\label{thm:Infeasible case}
 For a problem instance $\nu'$ that is $(x, \epsilon)$-infeasible, for any $(1-\delta)$-sound deterministic policy with $d=2$, $\delta < 1/2$,
    \begin{equation*}
      E_{\nu'}[\tau] \geq
    \sum_{i=1}^k \max \left\{ D(p_i|x-\epsilon)^{-1}, D(p_i|x+\epsilon)^{-1} \right\} 
    \frac{1}{2}
    \log(\frac{1}{4\delta}) 
    \end{equation*}
\end{theorem}
Lower bound proofs can be found in \cref{pf:Oracle Feasible case}.

%%%%%%%%%%%%%%%%%%%%%%%%%%%%%%%%%%%%%
\subsection{Sampling Policies}
\label{sec:sampling policies}
We present four sampling policies, a naive Uniform policy as a baseline along with Lower Upper Confidence Bound (LUCB) Mean, LUCB Ratio and Beta Thompson Sampling. We give high probability upper bounds for Uniform and LUCB Mean, and empirical evidence that LUCB Mean, LUCB Ratio, and Beta TS significantly outperform Uniform.

\textbf{Notation: }Let $B(n, \delta)$ be a confidence margin dependent upon sample size $n$ and confidence parameter $\delta$ such that 
\begin{equation}
\label{eq:margin}
    \sum_{n=1}^{\infty} P\left( |p_i - \hat{p}_i(n)| > B\left(n, \delta \right)\right) < \frac{\delta}{k}.
\end{equation}
    
We write $B_i(t)$ to represent the confidence margin for action $i$ given its sample size at time $t$ when $\delta$ is implied. Let $\hat{p}_i(t)$ be the estimated mean of action $i$ at time $t$, and $R_i(t) = \{y: \hat{p}_i(t) - B_i(t) \leq y \leq  \hat{p}_i(t) + B_i(t)\}$ be the confidence region of action $i$ at time $t$. We use $a_t$ to specify the action chosen at time $t$ and $n_i(t)$ the number of times action $i$ has been chosen at time $t$.
Each policy follows the same stopping rules for termination. 

We next define the direction of greatest uncertainty, which is used to determine termination and in our sampling policies for action selection. This measure aims to capture which direction away from $x$, we are least certain an action mean lies on. 

\begin{definition}[Direction of greatest uncertainty]
\label{def:uncertainty}
Given a confidence margin $B_i$ and mean estimate $\hat{p}_i$, the direction of greatest uncertainty $u \in \{1,-1\}$ is defined as,
\begin{equation*}
    u = \argmin_{u \in \{1,-1\}} \ \max_{i \in [k]} \  u(\hat{p}_i - x) - B_i.
\end{equation*}
\end{definition}

The intuition behind this definition is that it identifies the direction from $x$ we are furthest from determining a mean exists in that direction. For example, if $x=.5$, and there are two confidence regions $(.48, .9)$ and $(.49, .8)$, then the closest lower bound in direction $u=-1$ is .8, and the closest lower bound in direction $u=1$ is .48. The decision boundary that implies a mean lies below $x$ is further from $x$ than a decision boundary that implies a mean lies above it, so our direction of greatest uncertainty is $u=-1$ and we should sample actions that we have a higher belief are below $x$. 

All the policies presented follow the same stopping rules.

\textbf{Stopping Rules:} If one of the following criteria are met, the policy terminates,
\begin{enumerate}
    \item \textbf{Feasible:} $x_{\epsilon}$ is not separable from any subset consisting of a point from each of the confidence regions.\\
    $\min_{u \in \{-1,1\}} max_{i \in [k]} (\hat{p}_i -x) u - B_i(t) > - \epsilon$
    \label{sr:fB}
    \item \textbf{Infeasible:} $x_{\epsilon}$ is separable from all confidence regions.\\
    $\min_{u \in \{-1,1\}} max_{i \in [k]} (\hat{p}_i -x) u + B_i(t) < - \epsilon$
    \label{sr:ifB}
\end{enumerate}
Where stopping rule \ref{sr:fB} states there is a mean whose confidence interval lies above $x - \epsilon$ and one whose confidence intervals lies below $x + \epsilon$. The same confidence interval may satisfy both of these conditions. Intuitively, stopping rule \ref{sr:fB} says that if the true means lie in their respective confidence intervals, then no matter their value, a point in $x_{\epsilon}$ lies in their convex hull. 

\subsubsection{Uniform}
This simple policy samples from the active actions and chooses the action with the least samples, leading to uniform sample sizes across active actions. Active actions at time $t$ are those whose confidence regions at time $(t-1)$ contain a boundary point of $x_{\epsilon}$. The policy is given in \cref{alg:Uniform}.

\begin{algorithm}[ht]
    \DontPrintSemicolon
    \SetKwInOut{Input}{input}
    \SetKwInOut{Fix}{fix}
    \caption{Uniform Bernoulli}\label{alg:Uniform}
    \Input{Number of actions $k$, confidence $1-\delta$, $x$, $\epsilon$.}
    Sample from each source once.\;
    \While {Stop = False}{
        Update active actions $A_{t} = \{i : \exists y \in \partial x_{\epsilon}, y \in R_i(t)  \}$.\;
        $a_{t+1} = \argmin_{i \in A_{t}} n_i(t)$
        }
\end{algorithm}

%%%%%%%%%%%%%%%%%%%%%%%%%%%%%%%%%%%%%%
\subsubsection{LUCB Mean}

This policy is based on the idea of sampling the active action with the confidence boundary furthest from $x$ in the direction of greatest uncertainty, as given in \cref{def:uncertainty}. Given this direction, we exploit the action whose confidence bound is furthest from $x$. The policy is given in \cref{alg:LUCBM}.

\begin{algorithm}[ht]
    \DontPrintSemicolon
    \SetKwInOut{Input}{input}
    \SetKwInOut{Fix}{fix}
    \caption{LUCB Mean Bernoulli}\label{alg:LUCBM}
    \Input{Number of actions $k$, confidence $1-\delta$, $x$, $\epsilon$.}
    Sample from each source once.\;
    \While {Stop = False}{
        $u_t = \argmin_{u \in \{1, -1\}} \max_{i \in [k]} u(\hat{p}_i(t)
        - x) - B_i(t)$\;
        $a_{t+1} = \argmax_{i \in [k]} u_t(\hat{p}_i(t) - x) + B_i(t)$
        }
\end{algorithm}

%%%%%%%%%%%%%
\subsubsection{LUCB Ratio}
Using \cref{def:uncertainty} to define the direction of greatest uncertainty, the intuition of this policy is to sample from the active action whose confidence region has the largest proportion of area on the side of $x$ in this direction. It is possible that two actions have the same confidence ratio, at which point exploring the less sampled action provides more information. To account for this, we scale the confidence ratio by $\frac{1}{\sqrt{n_i}}$. The policy is given in \cref{alg:LUCBR}.

\begin{algorithm}[ht]
    \DontPrintSemicolon
    \SetKwInOut{Input}{input}
    \SetKwInOut{Fix}{fix}
    \caption{LUCB Ratio Bernoulli}\label{alg:LUCBR}
    \Input{Number of actions $k$, confidence $1-\delta$, $x$, $\epsilon$.}
    Sample from each source once.\;
    \While {Stop = False}{
         $u_t = \argmin_{u \in \{1, -1\}} \max_{i \in [k]} u(\hat{p}_i(t)
        - x) - B_i(t)$\;
        $a_{t+1} = \argmax_{i \in [k]} \frac{1}{\sqrt{n_i}}\frac{u_t(\hat{p}_i(t) - x) + B_i(t)}{u_t(x - \hat{p}_i(t))+ B_i(t)}$
        }
\end{algorithm}

%%%%%%%%%%%
\subsubsection{Thompson Sampling}
This probabilistic algorithm is a standard choice in the bandit literature. With few changes we adjust it to the convex hull feasibility problem. Again we use the direction of greatest uncertainty, sample a mean from the posterior of each action, and play the action with the mean furthest from $x$ in the given direction. The policy is given in \cref{alg:TS} where  $r_i(t)$ are the number of success drawn from action $i$ at time $t$.

\begin{algorithm}[ht]
\label{alg:TS}
    \DontPrintSemicolon
    \SetKwInOut{Input}{input}
    \SetKwInOut{Fix}{fix}
    \caption{Beta Thompson Sampling}
    \Input{Number of actions $k$, confidence $1-\delta$, $x$, $\epsilon$}
    Sample from each source once.\;
    \While {Stop = False}{
        Update posteriors $\pi_i(t) = Beta(1+ r_i(t), 1+ (n_i(t) - r_i(t)))$\;
       $u_t = \argmin_{u \in \{1, -1\}} \max_{i \in [k]} u(\hat{p}_i(t)
        - x) - B_i(t)$\;
        Sample $\tilde{p}_i(t)$ from posterior $\pi_i(t)$ for all $i \in [k]$\;
        $a_{t+1} = \argmax_{i \in [k]} u_{t}(\tilde{p}_i(t) - x)$ 
        }
\end{algorithm}

%%%%%%%%%%%%%%%%%%%%%%%%%%%%
\subsection{Sample Complexity Upper Bounds}
For both Uniform and LUCB Mean policies we give high probability upper bounds on the sample complexity. These policies sample all actions in relation to the optimal feasible subset, and thus allow a simple bounding on the complexity of each action. In \cref{sec:simulations}, we show that Beta TS, LUCB Ratio, and LUCB Mean outperform Uniform empirically.

\begin{figure}
    \centering
    \begin{tikzpicture}[ultra thick]
% \draw (1,1) node[anchor=east] {0} -- (5,1) node[anchor=north] {x} -- (10,1) node[anchor=west] {1};
\draw (0,1)  -- (9,1);
\draw (3,1) node {(};
\draw (7,1) node {)};
\filldraw (.5,1) circle[radius=1.5pt] node[below=5pt] {$p_i$};
\filldraw (5,1) circle[radius=1.5pt] node[below=5pt] {$x$};
% Calligraphic brace
\draw [pen colour={orange},
    decorate, 
    decoration = {calligraphic brace,
        raise=5pt,
        amplitude=10pt}] (.5,1) --  (7,1)
        node[pos=0.55,above=10pt,black]{$\Delta^{max}_i$};
\draw [pen colour={cyan},
    decorate, 
    decoration = {calligraphic brace, mirror,
        raise=5pt,
        amplitude=7pt}] (.5,1) --  (3,1)
        node[pos=0.6,below=10pt,black]{$\Delta^{min}_i$};
\draw [pen colour={black},
    decorate, 
    decoration = {calligraphic brace, mirror,
        raise=5pt,
        amplitude=3pt}] (5,1) --  (7,1)
        node[pos=0.5,below=6pt,black]{$\epsilon$};
\end{tikzpicture}
    \caption{Visualization of $\Delta_i^{max}, \Delta_i^{min}$ for some $p_i$ given $x, \epsilon$.}
    \label{fig:visual}
    % \Description[Visual relationship of gaps, means, and x]{Visual representation of notation. A point p lies outside the set x epsilon. the distance from point p to x, plus epsilon is delta max. The distance from p to x, minus epsilon, is delta min}
\end{figure}
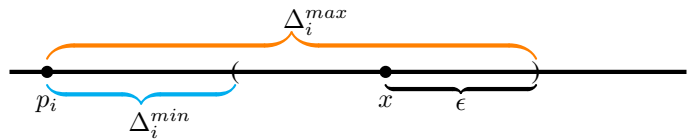

\textbf{Notation:}We define $\Delta^{max}_{i}$ to be the maximum distance from $p_i$ to a boundary of $x_{\epsilon}$ and define $\Delta^{min}_{i}$ to be the minimum distance from $p_{i}$ to a boundary of $x_{\epsilon}$. Let $s_i^{max}$ be the minimum integer solution to
$\Delta_i^{max} > 2B(s_i^{max}, \delta)$ and similarly for $s_i^{min}$. Therefore we have that with probability at least $1-\delta/k$, when action $i$ is sampled $s_i^{max} (s_i^{min})$ times, $\Delta_i^{max} (\Delta_i^{min})$ will not be contained in its confidence region. A visualization of the gap relationship is show in \cref{fig:visual}. We additionally define $\Delta_{i,j} = |p_i - p_j|$ and $s_{i,j}$ and the smallest integer such that $\Delta_{i,j} > 2B(s_{i,j})$. 

For the feasible Bernoulli setting, the optimal subset $\mathcal{J}^*$ will consist of one action, $\mathcal{J}^* = \{l^*\}$, or two actions, $\mathcal{J}^* = \{1,k\}$. In the following theorems we include the general case for any $B(n, \delta)$ that satisfies \cref{eq:margin} and where $s$ depends on choice of $B(n, \delta)$, as well as for the case with $B(n,\delta) = \sqrt{\frac{1}{2n}\log(n^2\frac{5k}{3\delta})}$, which shows the gap dependencies clearly.

\begin{theorem}[Uniform Complexity]
\label{thm:uniform}
Let $j^* = \argmax_{i \in \{1,k\}} s_i^{max}$. Assume $B(n, \delta)$ satisfies \cref{eq:margin}. When the underlying case is feasible, the sample complexity of Uniform is bounded above by

\begin{align*}
    &\tau \leq 
    \begin{cases}
    \mathcal{O} \left( \sum_{i=1}^k\min \left( s_{j^*}^{max}, s_i^{min} \right)  \right) & \mathcal{J}^* = \{1,k\}\\
    \mathcal{O} \left( \sum_{i=1}^k\min \left( s_{l^*}^{min}, s_i^{min} \right)  \right) & \mathcal{J}^* = \{l^*\}
    \end{cases}
\end{align*}
and for $B(n,\delta) = \sqrt{\frac{1}{2n}\log(n^2\frac{5k}{3\delta})}$,
\begin{align*}
    &\tau \leq 
    \begin{cases}
    \mathcal{O} \left( \sum_{i=1}^k\min \left( \frac{1}{(\Delta_{j^*}^{max})^2}, \frac{1}{(\Delta_{i}^{min})^2} \right)  \right) & \mathcal{J}^* = \{1,k\}\\
    \mathcal{O} \left( \sum_{i=1}^k\min \left( \frac{1}{(\Delta_{l^*}^{min})^2}, \frac{1}{(\Delta_{i}^{min})^2} \right)  \right) & \mathcal{J}^* = \{l^*\}
    \end{cases}
\end{align*}

with probability at least $1-\delta$.

When the cases is infeasible, the sample complexity of Uniform is bounded above by

\begin{align*}
    &\tau \leq \mathcal{O} \left(\sum_{i=1}^k s_i^{min}\right)
    &\tau \leq \mathcal{O} \left(\sum_{i=1}^k \frac{1}{(\Delta_{i}^{min})^2}\right)\\
    & &\text{For }B(n,\delta) = \sqrt{\frac{1}{2n}\log(n^2\frac{5k}{3\delta})}
\end{align*}

With probability at least $1-\delta$.
\end{theorem}

\begin{theorem}[LUCB Mean Complexity]
\label{thm:LUCBf}
Let $j^* = \argmax_{i \in \{1,k\}} s_i^{max}$ and $i^* = \argmin_{i \in \{1,k\}} s_i^{max}$. Assume $B(n, \delta)$ satisfies \cref{eq:margin}. When the underlying is feasible, the sample complexity of LUCB Mean is bounded above by

\[
    \tau \leq 
    \begin{cases}
    \mathcal{O} \left(\sum \limits_{i: \Delta_{i,j^*} \leq \Delta_{j^*}^{max}} s_{j^*}^{max} + \sum\limits_{i: \Delta_{i,j^*} > \Delta_{j^*}^{max}}  \max \left(s_{i,j^*}, s_{i^*}^{max}\right)\right)
    &\exists i, j, \ p_i < x < p_j 
    \\
    \mathcal{O} \left( \sum_{i=1}^k  \min \left(s_{i,j^*}, s_{j^*}^{min}\right)\right)
     & \text{otherwise}
    \end{cases}
\]
and  for $B(n,\delta) = \sqrt{\frac{1}{2n}\log(n^2\frac{5k}{3\delta})}$
\[
    \tau \leq 
    \begin{cases}
    \mathcal{O} \left(\sum \limits_{i: \Delta_{i,j^*} \leq \Delta_{j^*}^{max}} \frac{1}{(\Delta_{j^*}^{max})^2} + \sum \limits_{i: \Delta_{i,j^*} > \Delta_{j^*}^{max}} \max \left( \frac{1}{\Delta_{i,j^*}^2}, \frac{1}{(\Delta_{i^*}^{max})^2}\right)\right)
    &\exists i, j, \ p_i < x < p_j \\
     \mathcal{O} \left(\sum \limits_{i=1}^k \min \left( \frac{1}{\Delta_{i,j^*}^2}, \frac{1}{(\Delta_{j^*}^{min})^2}\right)\right)
     & \text{otherwise}
    \end{cases}
\]

with probability at last $1-
\delta$.

When the underlying case is infeasible, the sample complexity of LUCB Mean is bounded above by,

\begin{align*}
    &\tau \leq \mathcal{O} \left(\sum_{i=1}^k s_i^{min}\right)
    &\tau \leq \mathcal{O} \left(\sum_{i=1}^k \frac{1}{(\Delta_{i}^{min})^2}\right)\\
    & &\text{For }B(n,\delta) = \sqrt{\frac{1}{2n}\log(n^2\frac{5k}{3\delta})}
\end{align*}

With probability at least $1-\delta$.
\end{theorem}

This shows that for any problem instance, the worst case sample complexity is lower using the LUCB Mean policy compared to the Uniform policy, since $\min (s_{j^*}^{max}, s_l^{min}) \geq \max (s_{l,j^*}, s_{i^*}^{max})$ for all $l \in [k]$. We leave details of this to \cref{ap:LUCB-U}.

Intuitively, \cref{thm:uniform} says that all actions are sampled as many times as the most sampled optimal arm or until it's confidence region is disjoint from $x_{\epsilon}$. For \cref{thm:LUCBf}, the bounds are more complicated because it depends upon how close the action mean is to $p_{j^*}$ and the instance setting. Generally speaking, the bounds describe a relationship between the relative distance of an action's mean and the most sampled optimal action's mean, an action's mean and a boundary point in $x_{\epsilon}$, and the worst case behavior of the action's confidence region. These particular's are detailed in the proof.

%%%%%%%%%%%%%%%%%%%%%%%%%%%%%%%%%%%%%
\section{Multinomial Feasibility Sampling}
By expanding our definition of the direction of greatest uncertainty and our stopping rules, we can modify each policy to work in higher dimensions. 

\subsection{Feasibility and Infeasibility Checks}
Recall the definition of $1-\delta$ Confident Feasible (\cref{def:df}). If we assume $\epsilon = 0$, an equivalent definition would be
\begin{equation*}
    \forall u, ||u||=1, \ \exists R_i \text{ such that }  \left(q_i - \left(x+w\right)\right)^T u > 0 \ \forall q_i \in R_i
\end{equation*}
which states that $x$ is not separable from any subset of points constructed from the confidence regions. Alternatively, we may say that for all unit vectors $u$, $\max_{i \in [k]} \min_{q \in R_i} \ (q_i - x)^Tu > 0$. If we limit the confidence regions to be balls with radius $B$, then we can simplify to say $x$ is feasible if 
\begin{equation*}
   \min_{u: ||u||=1} \max_{i \in [k]} (\hat{p}_i - x)^Tu - B_i > 0.
\end{equation*}

Now considering $\epsilon > 0$, we would need to show there exists some point $(x+w) \in x_ \epsilon$, $||w|| < \epsilon$,
\begin{equation*}
    \min_{u: ||u||=1} \max_{i \in [k]} (\hat{p}_i - (x+w))^Tu -B_i > 0
\end{equation*}
We have that for some $\lambda \in (0,1)$,
\begin{align*}
    &\min_{u: ||u||=1} \max_{i \in [k]} (\hat{p}_i - x)^Tu - B_i > -\lambda \epsilon\\
    &\min_{u: ||u||=1} \max_{i \in [k]} (\hat{p}_i - x)^Tu - B_i - w^Tu > -\lambda \epsilon - w^Tu\\
    &\min_{u: ||u||=1} \max_{i \in [k]} (\hat{p}_i - (x+w))^Tu - B_i - w^Ta > 0 
\end{align*}
Where in the last line we have that since $u$ is a unit vector and the length of $w$ is bounded by $\epsilon$, we can pick $w$, $\lambda \in (0,1)$ such that $w^Tu = -\lambda \epsilon$. Therefore a feasibility check becomes if, given some $\lambda \in (0,1)$,
\begin{equation*}
    \min_{u: ||u||=1} \max_{i \in [k]} (\hat{p}_i - (x+w))^Tu -B_i > 0.
\end{equation*}
In a similar fashion, an infeasibility check would be if there exists a unit vector $a$ such that,
\begin{equation*}
    \min_{u: ||u||=1} \max_{i \in [k]} (\hat{p}_i - x)^Tu + B_i < -\epsilon.
\end{equation*}

%%%%%%%%%%%%%%%%%%%%%%%%%%%%%%%%%%%%%
\subsection{Sampling Policies}
Using the above formulation for checking feasibility lends itself to defining the direction of greatest uncertainty in any dimension. 

\begin{definition}[Direction of greatest uncertainty]
Given a confidence margin $B_i$ and mean estimate $\hat{p}_i$, the direction of greatest uncertainty $a$ is defined as,
\begin{equation*}
    u = \argmin_{u: \ ||u||=1} \max_{i \in [k]} (\hat{p}_i - x)^T u - B_i.
\end{equation*}
\end{definition}

Unfortunately, finding the direction of greatest uncertainty for $d \geq 3$, and thus also checking feasibility, is a non-convex problem, so we cannot obtain the optimal solution. One obvious workaround to this is simply doing a grid search over some subset of points on the unit ball. This is the approach we take.

Let $G$ be some subset of the unit ball in dimension $d$ which will be the directions we search over, and let $\lambda \in (0,1)$ be a parameter. 

\textbf{Stopping Rules:} If one of the following criteria are met, the policy terminates,
\begin{enumerate}
    \item $x_\epsilon$ is not separable from the confidence balls in any direction $u \in G$.\\
        $\min_{u \in G} \max_{i \in [k]} (\hat{p}_i(t) - x)^T u - B_i(t) > - \lambda \epsilon$
    \item $x_{\epsilon}$ is separable from all confidence balls.\\
         $\min_{u \in G} \max_{i \in [k]} (\hat{p}_i(t) - x)^T u + B_i(t) < - \epsilon$
\end{enumerate}

Our sampling algorithms do not change significantly to accommodate higher dimensions. The Uniform policy no longer has an active action set, and the other policies use the updated definition of direction of greatest uncertainty and vector dot products instead of scalar multiplication. The policies are given in \cref{alg:UniformH} (Uniform), \cref{alg:LUCBH} (LUCB Mean), \cref{alg:CRH} (LUCB Ratio), and \cref{alg:TSH} (Dirichlet Thompson sampling).

\begin{algorithm}[ht]
    \DontPrintSemicolon
    \SetKwInOut{Input}{input}
    \SetKwInOut{Fix}{fix}
    \caption{Uniform}\label{alg:UniformH}
    \Input{Number of actions $k$, confidence $1-\delta$, $x$, $\epsilon$.}
    Sample from each source once.\;
    \While {Stop = False}{
        $a_{t+1} = \argmin_{i \in [k]} n_i(t)$
        }
\end{algorithm}

\begin{algorithm}[ht]
    \DontPrintSemicolon
    \SetKwInOut{Input}{input}
    \SetKwInOut{Fix}{fix}
    \caption{LUCB Sampling}\label{alg:LUCBH}
    \Input{Number of actions $K$, confidence $1-\delta$, unit vectors $G$.}
    \Fix{$A = [k]$}
    Sample from each source once.\;
    \While {Stop = False}{
        $u_t = \argmin_{u \in G} umax_{i \in [k]} (\hat{p}_i - x)^T u -B_i(t)$\\
        $a_{t+1} = \argmax_{i \in [k]} (\hat{p}_i(t)-x)^T u_t + B_i(t)$
        }
\end{algorithm}

\begin{algorithm}[ht]
    \DontPrintSemicolon
    \SetKwInOut{Input}{input}
    \SetKwInOut{Fix}{fix}
    \caption{Confidence Ratio Sampling}\label{alg:CRH}
    \Input{Number of actions $K$, confidence $1-\delta$, unit vectors $G$.}
    Sample from each source once.\;
    \While {Stop = False}{
        $u_t = \argmin_{u \in G} \max_{i \in [k]} (\hat{p}_i - x)^T u -B_i(t)$\\
        $a_{t+1} = \argmax_{i \in [k]} \frac{1}{\sqrt{n_i}}\frac{(\hat{p}_i(t) - x)^T u_t + B_i(t)}{(x - \hat{p}_i(t))^T u_t + B_i(t)}$
        }
\end{algorithm}

\begin{algorithm}[ht]
    \DontPrintSemicolon
    \SetKwInOut{Input}{input}
    \SetKwInOut{Fix}{fix}
    \caption{Dirichlet Thompson Sampling}\label{alg:TSH}
    \Input{Number of actions $K$, confidence $1-\delta$, priors $\pi_i$, unit vectors $G$}
    Sample from each source once.\;
    \While {Stop = False}{
         $u_t = \argmin_{a \in G} \max_{i \in [k]} (\hat{p}_i - x)^T u -B_i(t)$\\
        Sample $p_i(t)$ from posterior $\pi_i(t)$ for all $i \in [k]$\;
        $a_{t+1} = \argmax_{i \in [k]} (p_i(t) - x)^T u_{t}$ 
        }
    Solve for $t$ in plug optimization. If $t \geq 0$, out = \emph{feasible}, else if $t < 0$ out = \emph{infeasible}.
\end{algorithm}

%%%%%%%%%%%%%%%%%%%%%%%%%%%%%%%%%%%%%
\section{Simulations}
\label{sec:simulations}
We compare the average sample size till termination of our three policies against the naive uniform sampling method.

\subsection{Setup}
We run our policies in the Bernoulli setting (which correlates to both $d=1$ and $d=2$) and the Multinomial setting with $d=3$. Each graph shows the average sample size at termination for the four policies when averaged over 30 trials using $B(n,\delta) = \sqrt{\frac{1}{2n}\log(n^2\frac{5k}{3\delta})}$. In all trials, we set $\delta = .01$, $k=10$, $\epsilon= 0.1$, and set $\lambda=.99$ when $d=3$. In the Multinomial setting, we use a grid search over 300 points on the unit sphere. For the Bernoulli setting we run scenarios for $|\mathcal{J}^*| \in \{1,2\}$ and for the Multinomial setting $|\mathcal{J}^*| \in \{1,2, 3\}$. In each of these settings we further consider two cases, when $\mathcal{J}^*$ is unique and when it is not. When the $\mathcal{J}^*$ is unique, it is an element of the set of optimal subsets when $\mathcal{J}^*$ is not unique. Therefore the oracle lower bound is the same for unique and non-unique cases and we can compare the two outcomes when all other parameters are fixed. The setting for each scenario is listed in the caption. The desired values is in the Bernoulli case $x=.5$, and $x = (.33, .33, .33)$ in the Multinomial case. The means used in each setting are listed in \cref{tab:bernoulli means,tab:multinomial means}, and were chosen as a general representation of several different scenarios.

\begin{table}[ht]
\caption{Bernoulli Mean Values}
\label{tab:bernoulli means}
\centering
\begin{tabular}{l|l|l}
\toprule
 Bernoulli & $|\mathcal{J}^*|=1$ & $|J^*|=2$  \\
 \midrule
 Optimal  & .5 & .3, .7   \\
 \midrule
Non-optimal & .48, .52 & .48, .52\\
\bottomrule
\end{tabular}
\end{table}

\begin{table}[ht]
\caption{Multinomial Mean Vectors}
\label{tab:multinomial means}
\centering
\begin{tabular}{l|l|l|l}
\toprule
 Multinomial & $|J^*|=1$ & $|J^*|=2$ &  $|J^*|=3$\\
 \midrule
 Optimal  & (.33, .33, .33) & \makecell[l]{(.1, .57, .33)\\ (.57, .1, .33)}  & \makecell[l]{(.2, .1, .7)\\ (.7, .2, .1) \\ (.1, .7, .2)} \\
 \midrule
Non-optimal &  
(0, 0, .1) & \makecell[l]{(.2, .47, .33)\\ (.47, .2, .33)} & \makecell[l]{(.33, .33, .34)\\ (.33, .34, .33) \\ (.34, .33, .33)}\\
\bottomrule
\end{tabular}
\end{table}

\subsection{Results}
When the average sample size of the Uniform policy is substantially larger than that of the best performing policy, the y-axis has a break point to indicate a change in the scale. 

\begin{figure}
\centering
   \begin{subfigure}{.4\textwidth}
    \centering
     \includegraphics[trim={0  0 2cm 1cm},clip,width=\textwidth]{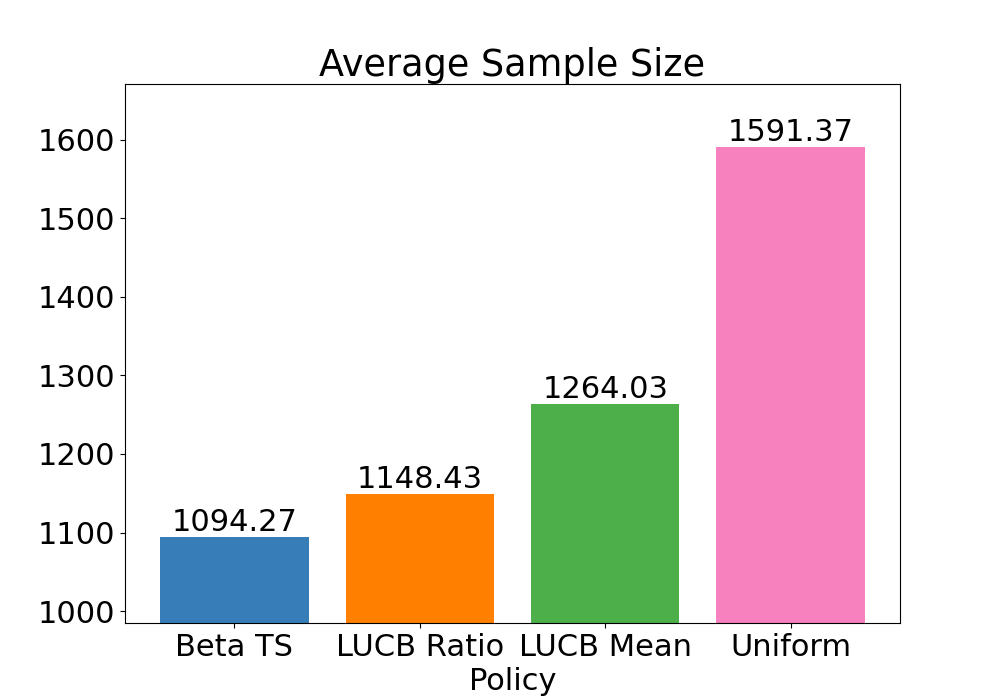}
    \caption{$\mathcal{J}^*$ unique.}% \caption{$d=2,  \ |\mathcal{J}^*| = 1, \ p = \{.3, .5^4, .6^4, .7\}$}
    \label{subfig:2d_one_diff}
  \end{subfigure}
  \begin{subfigure}{.4\textwidth}
    \centering
     \includegraphics[trim={0  0 2cm 1cm},clip,width=\textwidth]{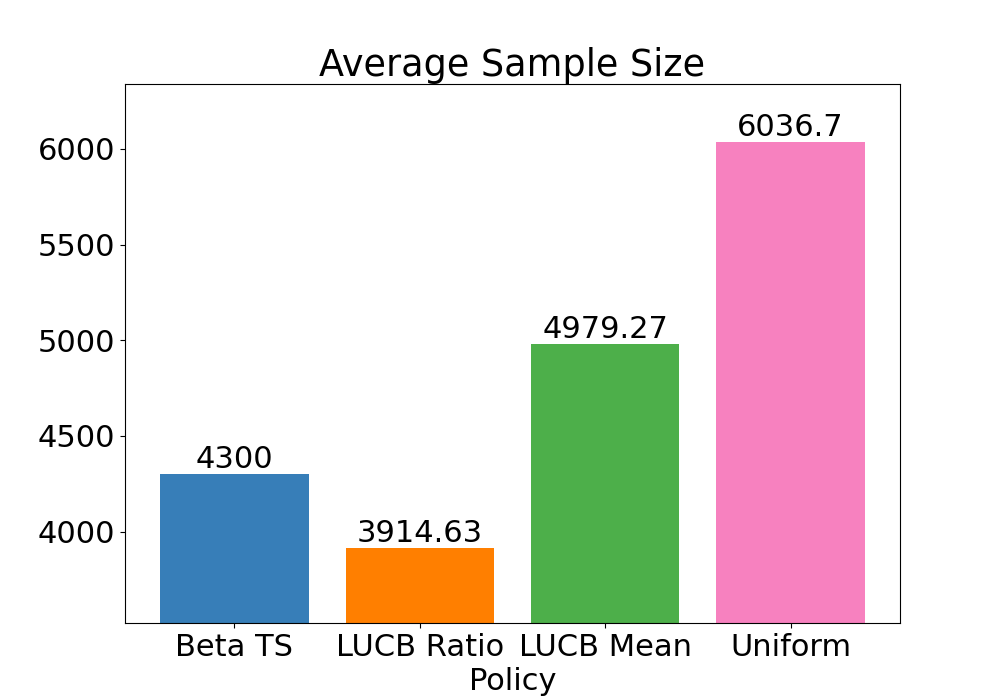}
    \caption{$\mathcal{J}^*$ not unique.}
    % \caption{$d=1,  \ |\mathcal{J}^*| = 1,\ p = \{.7^{10}\}$}
    \label{subfig:2d_one_equal}
  \end{subfigure}
  \caption{Average stopping time in Bernoulli setting,  $d=1, \ |\mathcal{J}^*| = 1$, $k=10$.}
  \label{fig:2d_one}
%   \Description[Beta T S optimal when J start unique, L U C B ratio optimal when J star not unique]{When J star is unique, the average sample sizes were 1094 for beta t s, 1148 for l u c b ratio, 1264 fr l u c b mean, and 1591 for uniform. When j star is not unique, beta t s is 4300, l u c b ratio is 3914, l u c b mean is 4927, and uniform is 6026}
\end{figure}

\begin{figure}
\centering
  \begin{subfigure}{.4\textwidth}
    \centering
     \includegraphics[trim={0  0 2cm 1cm},width=\textwidth]{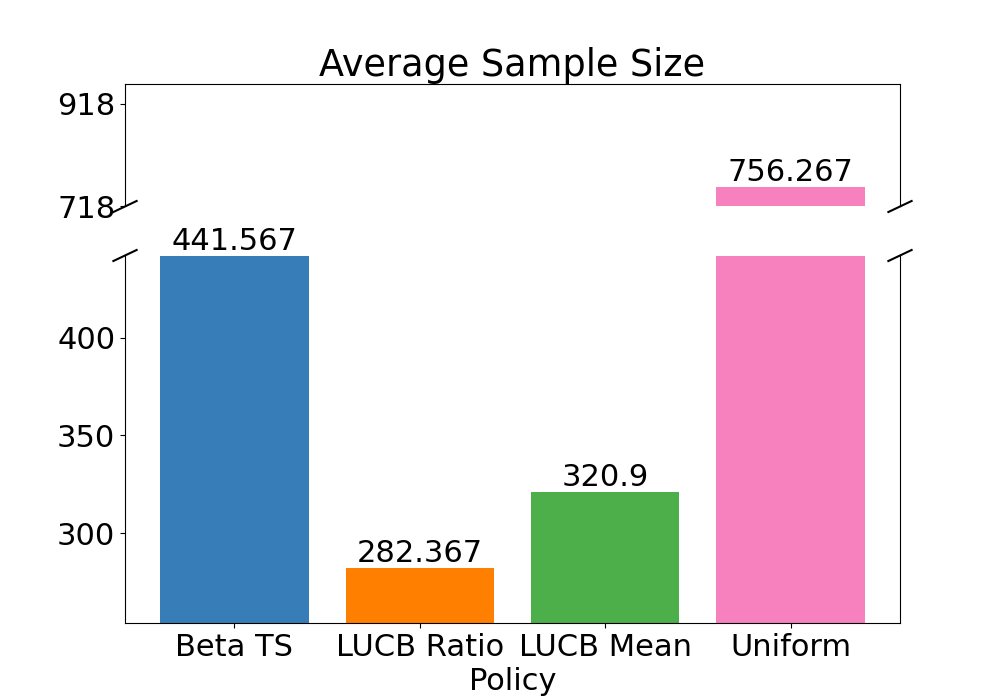}
     \caption{$\mathcal{J}^*$ unique.}
    % \caption{$d=2,  \ |\mathcal{J}^*| = 2,\ p = \{.3, .49^4, .51^4, .7\}$}
    \label{subfig:2d_two_diff}
  \end{subfigure}
  \begin{subfigure}{.4\textwidth}
    \centering
     \includegraphics[trim={0  0 2cm 1cm},width=\textwidth]{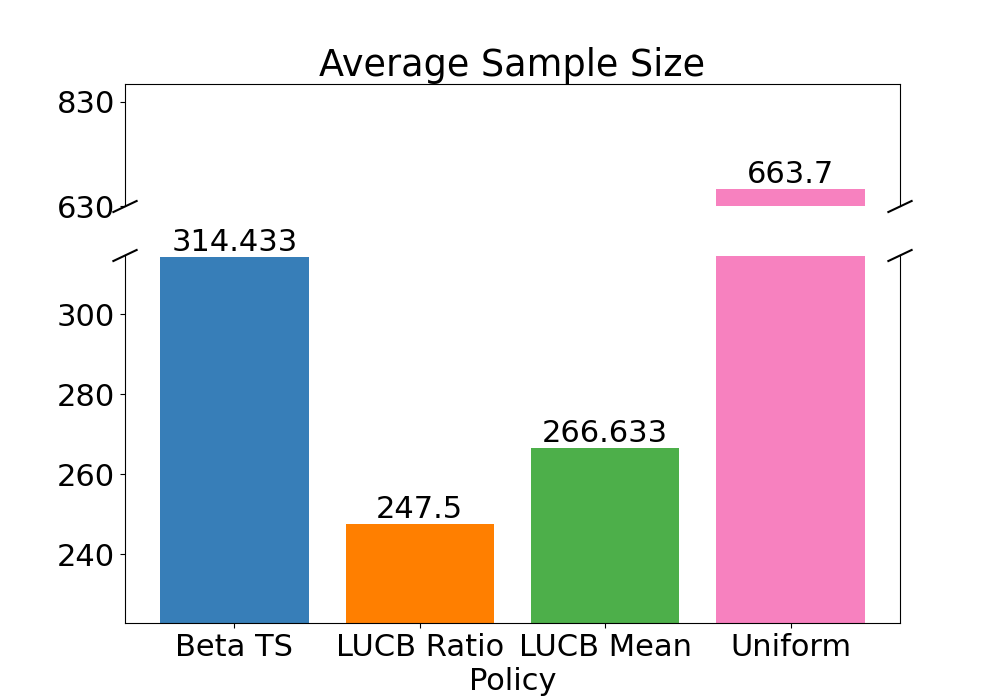}
    \caption{$\mathcal{J}^*$ not unique.}
    % \caption{$d=2,  \ |\mathcal{J}^*| = 2,\ p = \{.3^5, .7^5\}$}
    \label{subfig:2d_two_equal}
  \end{subfigure}
  \caption{Average stopping time in Bernoulli setting,  $d=1, \ |\mathcal{J}^*| = 2$, $k=10$.}
  \label{fig:2d_two}
%   \Description[L U C B ratio optimal in both case]{When J star is unique, the average sample sizes were 441 for beta t s, 282 for l u c b ratio, 320 fr l u c b mean, and 756 for uniform. When j star is not unique, beta t s is 314, l u c b ratio is 247, l u c b mean is 266, and uniform is 663}
\end{figure}

\begin{figure}
\centering
  \begin{subfigure}{.4\textwidth}
    \centering
     \includegraphics[trim={0  0 2cm 1cm},clip,width=\textwidth]{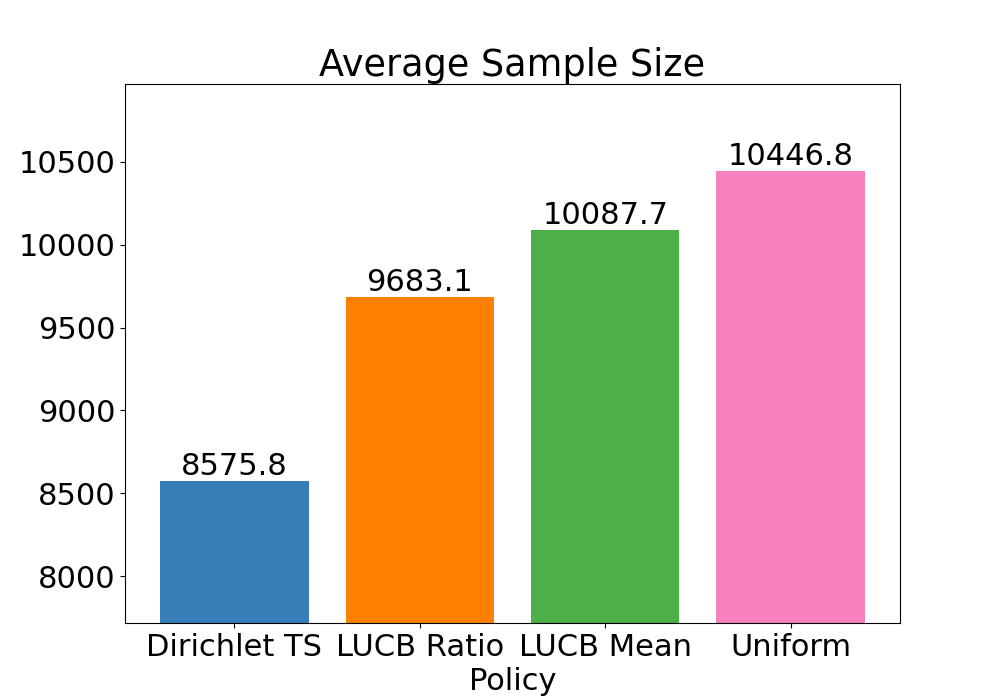}
    \caption{$\mathcal{J}^*$ unique.}
    \label{subfig:3d_one_diff}
  \end{subfigure}
  \begin{subfigure}{.4\textwidth}
    \centering
     \includegraphics[trim={0  0 2cm 1cm},clip,width=\textwidth]{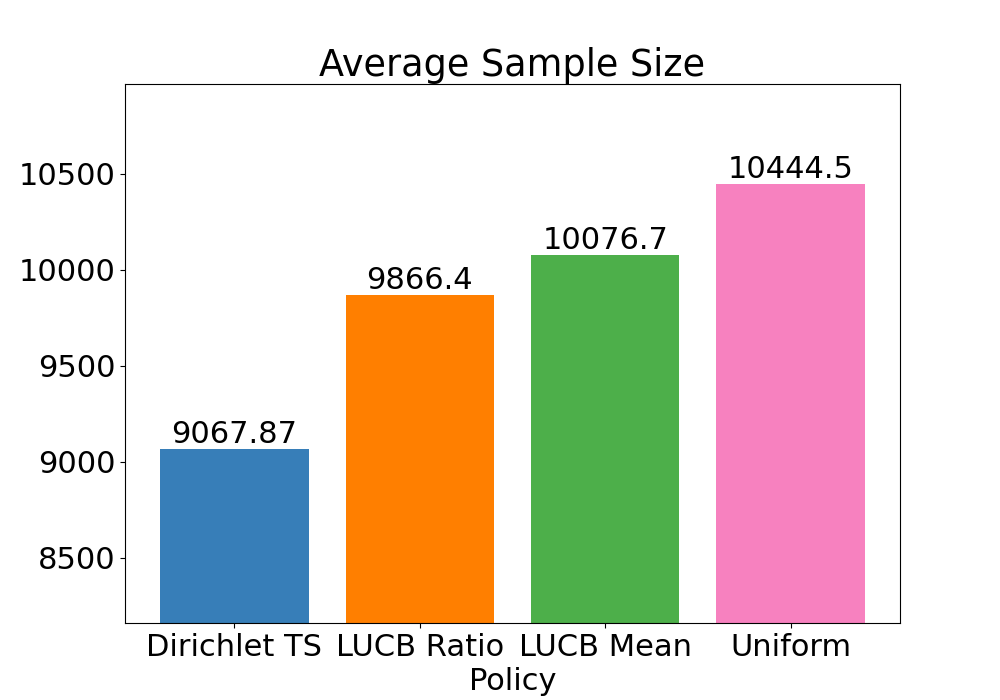}
    \caption{$\mathcal{J}^*$ not unique.}
    \label{subfig:3d_one_equal}
  \end{subfigure}
  \caption{Average stopping time in Multinomial setting, $d=3,  \ |\mathcal{J}^*| = 1$, $k=10$.}
  \label{fig:3d_one}
%   \Description[Beta T S optimal in both cases]{When J star is unique, the average sample sizes were 4901 for beta t s, 7761 for l u c b ratio, 7581 fr l u c b mean, and 13428 for uniform. When j star is not unique, beta t s is 9067, l u c b ratio is 9866, l u c b mean is 10076, and uniform is 10444}
\end{figure}

\begin{figure}
\centering
  \begin{subfigure}{.4\textwidth}
    \centering
     \includegraphics[trim={0  0 2cm 1cm},clip,width=\textwidth]{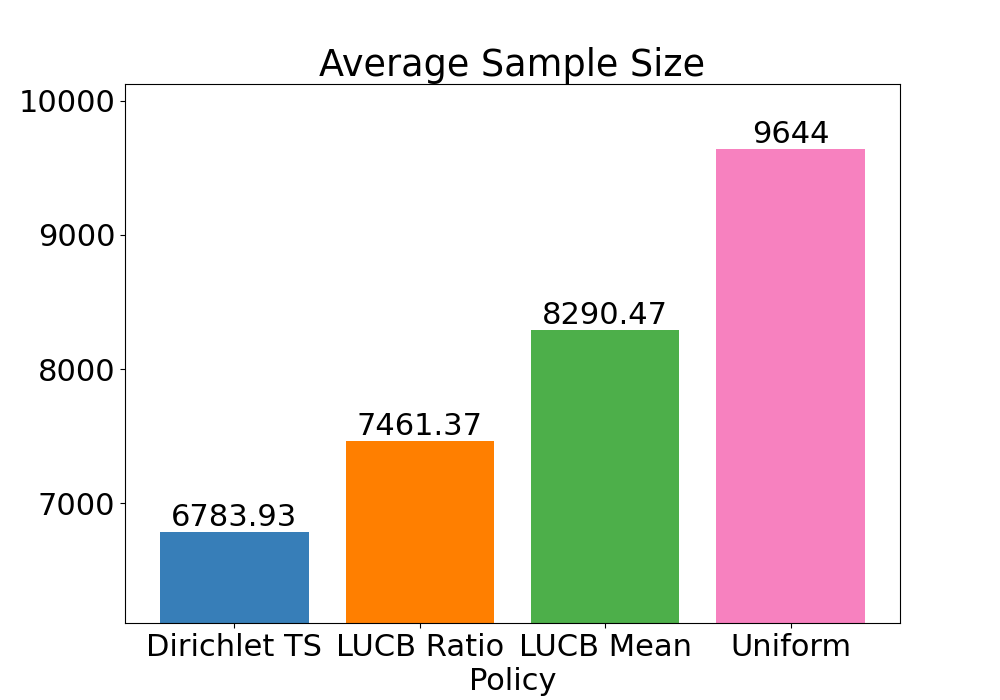}
    \caption{$\mathcal{J}^*$ unique.}
    \label{subfig:3d_two_diff}
  \end{subfigure}
  \begin{subfigure}{.4\textwidth}
    \centering
     \includegraphics[trim={0  0 2cm 1cm},clip,width=\textwidth]{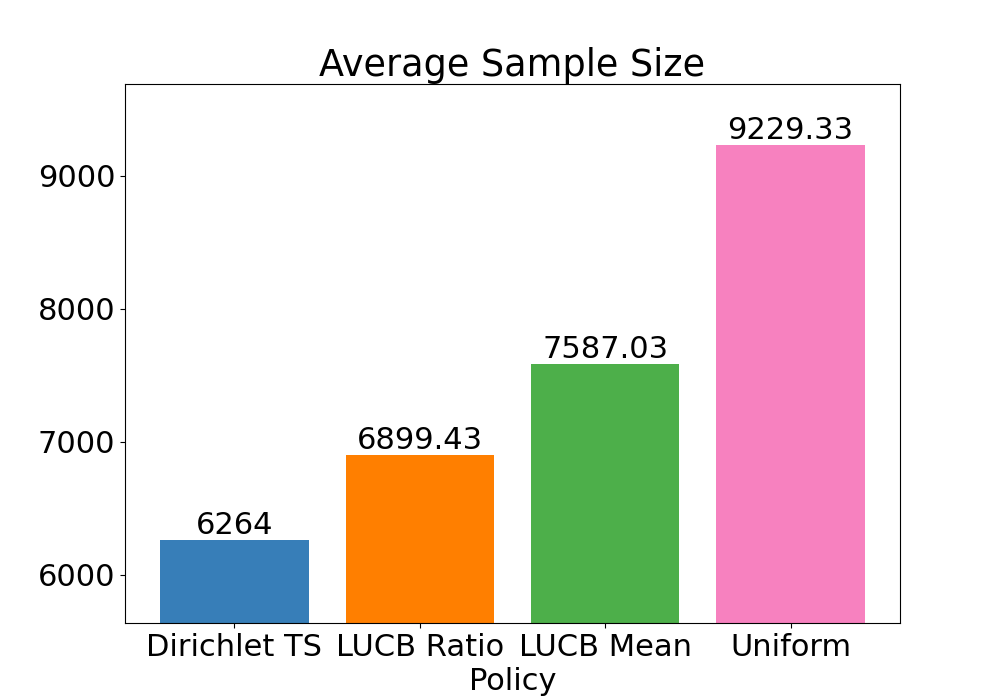}
    \caption{$\mathcal{J}^*$ not unique.}
    \label{subfig:3d_two_equal}
  \end{subfigure}
  \caption{Average stopping time in Multinomial setting, $d=3,  \ |\mathcal{J}^*| = 2$, $k=10$.}
  \label{fig:3d_two}
%   \Description[Beta T S optimal in both cases]{When J star is unique, the average sample sizes were 2866 for beta t s, 3513 for l u c b ratio, 4006 fr l u c b mean, and 11230 for uniform. When j star is not unique, beta t s is 6264, l u c b ratio is 6899, l u c b mean is 7587, and uniform is 9229}
\end{figure}

\begin{figure}
\centering
    \begin{subfigure}{.4\textwidth}
    \centering
     \includegraphics[trim={0  0 2cm 1cm},width=\textwidth]{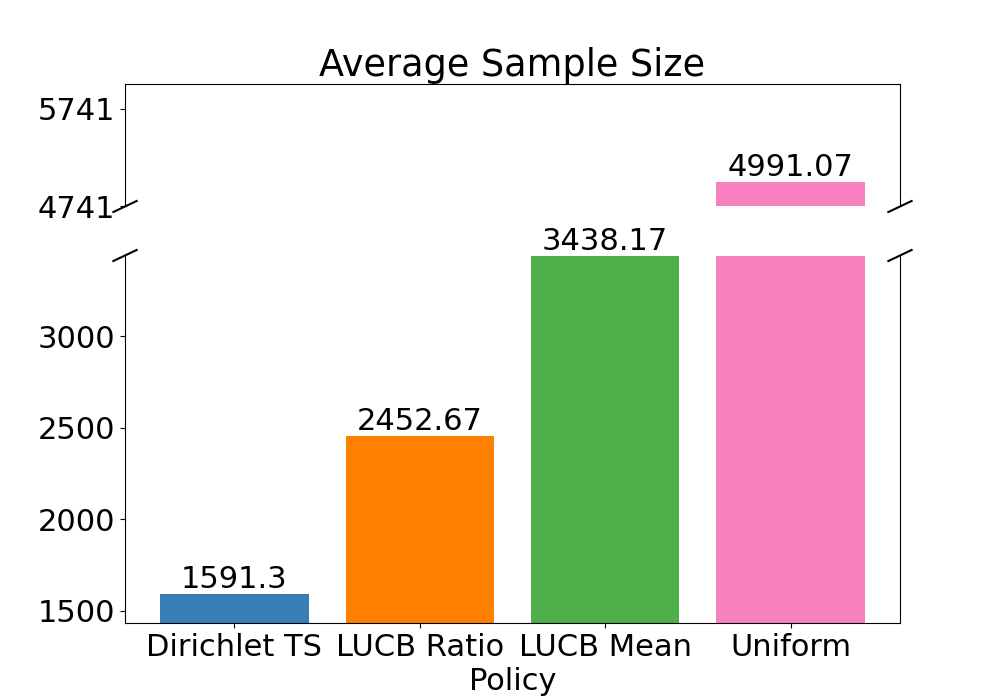}
    \caption{$\mathcal{J}^*$ unique.}
    \label{subfig:3d_three_diff}
  \end{subfigure}
  \begin{subfigure}{.4\textwidth}
    \centering
     \includegraphics[trim={0  0 2cm 1cm},width=\textwidth]{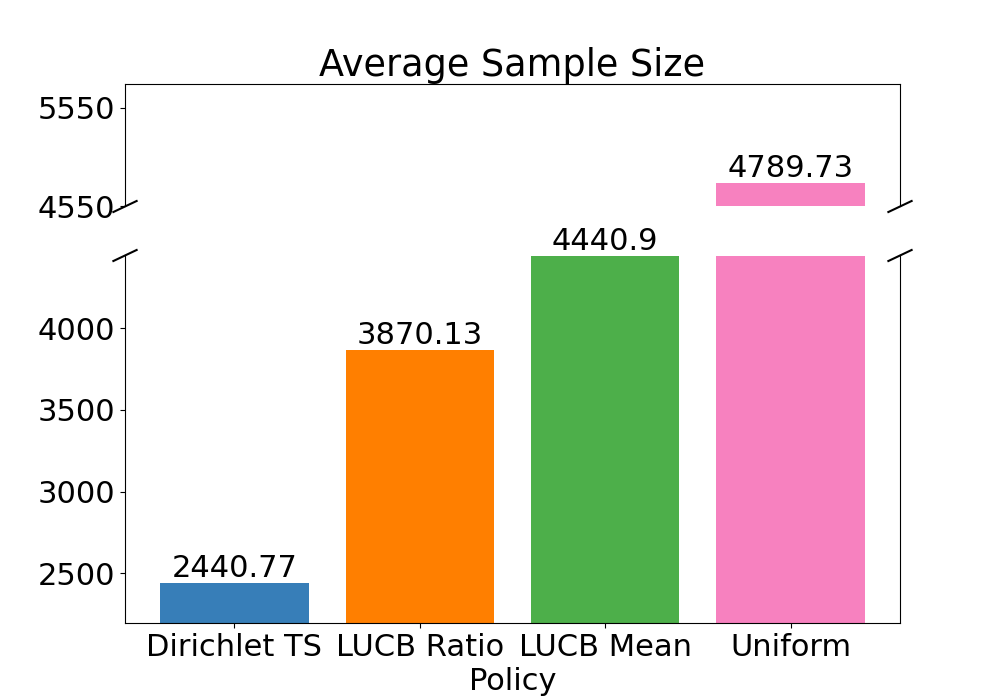}
    \caption{$\mathcal{J}^*$ not unique.}
    \label{subfig:3d_three_equal}
  \end{subfigure}
  \caption{Average stopping time in Multinomial setting, $d=3,  \ |\mathcal{J}^*| = 3$, $k=10$.}
  \label{fig:3d_three}
%   \Description[Beta T S optimal in both cases]{When J star is unique, the average sample sizes were 1591 for beta t s, 2452 for l u c b ratio, 3438 fr l u c b mean, and 4991 for uniform. When j star is not unique, beta t s is 2440, l u c b ratio is 3870, l u c b mean is 4440, and uniform is 4789}
  \end{figure}

\Cref{fig:2d_one,fig:2d_two} show results for Bernoulli sampling and \cref{fig:3d_one,fig:3d_two,fig:3d_three} show resuts for the Multinomial sampling with $d=3$. It is clear that the Uniform sampling policy performs the worst in all cases, and is improved upon by all other policies presented in this paper. It is clearly seen, and somewhat surprising, that there is a large relative difference in performance of LUCB Ratio and Thompson sampling between the Bernoulli and Multinomial setting. 

In the Multinomial setting Dirichlet Thompson sampling has superior performance, while in the Bernoulli setting LUCB Ratio has the best performance, except when there in one unique optimal action, as seen in \cref{subfig:2d_one_diff}. Here Beta Thompson sampling (Beta TS) outperforms the other policies. We speculate that in this particular Bernoulli setting, Beta TS this may be because this case is most similar to the standard multi-armed bandit problem, which aims to select the action with the highest mean as often as possible. The multi-armed bandit Beta TS policy is one of the simplest and most effective policies in practice  \citep{chapelleEmpiricalEvaluationThompson2011}. 

Looking at \cref{fig:2d_one}, when $\mathcal{J}^*$ is unique it requires fewer sample sizes on average for each policy than when $\mathcal{J}^*$ is not unique. This relationship reversed in \cref{fig:2d_two}. This example shows that uniqueness of $\mathcal{J}^*$ in the Bernoulli setting does not imply a simpler problem. This is similarly seen in the Multinomial setting. We see in \cref{fig:3d_one} that Dirichlet TS and LUCB Ratio perform better in the unique optimal subset setting, and there is no difference for LUCB Mean and Uniform. Whereas in \cref{fig:3d_two}, all but LUCB Ratio perform better in the non-unique optimal subset setting.

We see in both the Bernoulli and Multinomial setting that the larger the optimal subset, the fewer average samples before termination. This is because when $|\mathcal{J}^*| < d$ the optimal subsets must be sampled until $B(n,\delta) \approx \epsilon$ to ensure there is a mean either on both sides of $x$ or that a confidence region is fully contained in $x_{\epsilon}$ in all directions. 

In practice, results will be dependent upon the underlying truth, as can be inferred by the Oracle average sample complexity lower bound and the high probably sample complexity upper bounds given in this work. These simulations give evidence of the magnitude of improvement using an adaptive sampling method over the naive uniform method. Depending on the setting, average sample size can be reasonably small, as seen in \cref{subfig:2d_two_equal,subfig:2d_two_diff}. In the Multinomial setting, average sample sizes are in the thousands. The practicality of this method can be seen to depend upon the true distributions, sampling budget, and parameter values.

%%%%%%%%%%%%%%%%%%%%%%%%%%%%%%%%%%%%%
\section{Summary and Discussion}
We introduce the convex hull feasibility problem in the context of fair data collection. In the Bernoulli setting, we give a lower bound on the expected sample complexity in the $(x, \epsilon)$-infeasible instance and an oracle lower bound on the expected sample complexity in the $(x, \epsilon)$-feasible instance. We introduce four sampling policies for the Bernoulli setting, Uniform, LUCB Mean, LUCB Ratio, and Beta TS and give high probability upper bounds on sample complexity for the Uniform and LUCB Mean policies. We give the adaptation of the Binomial policies to the Multinomial case. Through simulation, we show LUCB Mean, LUCB Ratio, and the Thompson sampling policies significantly outperform Uniform in the Bernoulli and Multinomial setting. Under our simulation scenarios, we see that LUCB Ratio is typically the best performing policy in our Bernoulli settings, while Dirichlet TS is the best performing policy in our Multinomial settings. We discuss that the practicality of implementation is dependent upon the underlying distributions, sampling budget, and chosen parameters. Large sampling budgets would enable this method practical under most settings, whereas with small sampling budgets this method would only be practical if there was a strong prior that the underlying distributions has a small oracle lower bound.

While this work focused on Bernoulli and Multinomial convex hull feasibility sampling, the general problem is applicable when points are drawn from any distribution for which one can construct a confidence region that satisfies \cref{eq:margin}. 

There are some limitations within this work. Notably, we were only able to give an oracle lower bound on the expected sample complexity in the feasible case. A true lower bound would allow for better comparison of a policy's theoretical performance. Addtionally, we provide theoretical results in the Bernoulli settings, but not the Multinomial setting. 

The work in this paper is somewhat analogous to multi-armed bandit best arm identification with fixed-confidence. Another approach seen in the best arm identification literature is the fixed-budget setting, which could also be applied to the convex hull feasibility problem. If given a set of samples, the confidence regions can be such that they do not meet the definition of either $(1-\delta)$-confident feasible or $(1-\delta)$-confident infeasible. In this case we could ask instead what is the probability of feasibility or infeasible given the current sample, or if adaptively sampling, what is the highest probably of a correct decision when sampling with a budget. One application of this could be to check Pareto frontier feasibility, where if given noisy gradients, we ask what is the probability all groups can be improved versus the probability that improving some groups may harm others.

\newpage
\printbibliography

@inproceedings{abernethyAdaptiveSamplingReduce2020a,
  title = {Adaptive {{Sampling}} to {{Reduce Disparate Performance}}},
  booktitle = {{{arXiv}}:2006.06879 [Cs, Stat]},
  author = {Abernethy, Jacob and Awasthi, Pranjal and Kleindessner, Matth{\"a}us and Morgenstern, Jamie and Zhang, Jie},
  year = {2020},
  month = jun,
  eprint = {2006.06879},
  eprinttype = {arxiv},
  primaryclass = {cs, stat},
  archiveprefix = {arXiv}
}

@inproceedings{anahidehFairActiveLearning2021,
  title = {Fair {{Active Learning}}},
  booktitle = {{{arXiv}}:2001.01796 [Cs, Stat]},
  author = {Anahideh, Hadis and Asudeh, Abolfazl and Thirumuruganathan, Saravanan},
  year = {2021},
  month = mar,
  eprint = {2001.01796},
  eprinttype = {arxiv},
  primaryclass = {cs, stat},
  archiveprefix = {arXiv}
}

@inproceedings{asudehAssessingRemedyingCoverage2019,
  title = {Assessing and {{Remedying Coverage}} for a {{Given Dataset}}},
  booktitle = {International {{Conference}} on {{Data Engineering}}},
  author = {Asudeh, Abolfazl and Jin, Zhongjun and Jagadish, H. V.},
  year = {2019},
  month = feb,
  eprint = {1810.06742},
  eprinttype = {arxiv},
  publisher = {{IEEE}},
  archiveprefix = {arXiv}
}

@inproceedings{chapelleEmpiricalEvaluationThompson2011,
  title = {An {{Empirical Evaluation}} of {{Thompson Sampling}}},
  booktitle = {Advances in {{Neural Information Processing Systems}}},
  author = {Chapelle, Olivier and Li, Lihong},
  year = {2011},
  volume = {24},
  publisher = {{Curran Associates, Inc.}}
}

@inproceedings{dworkFairnessAwareness2012,
  title = {Fairness {{Through Awareness}}},
  booktitle = {Proceedings of the 3rd Innovations in Theoretical Computer Science Conference},
  author = {Dwork, Cynthia and Hardt, Moritz and Pitassi, Toniann and Reingold, Omer and Zemel, Rich},
  year = {2012},
  eprint = {1104.3913},
  eprinttype = {arxiv},
  archiveprefix = {arXiv},
  langid = {english}
}

@inproceedings{even-darPACBoundsMultiarmed2002,
  title = {{{PAC}} Bounds for Multi-Armed Bandit and {{Markov}} Decision Processes},
  booktitle = {In {{Fifteenth Annual Conference}} on {{Computational Learning Theory}} ({{COLT}})},
  author = {{Even-dar}, Eyal and Mannor, Shie and Mansour, Yishay},
  year = {2002},
  pages = {255--270}
}

@article{finkHyperplaneSeparabilityConvexity2017,
  title = {Hyperplane Separability and Convexity of Probabilistic Point Sets},
  author = {Fink, Martin and Hershberger, John and Kumar, Nirman and Suri, Subhash},
  year = {2017},
  month = feb,
  journal = {Journal of Computational Geometry (Old Web Site)},
  volume = {8},
  number = {2},
  pages = {32--57},
  issn = {1920-180X},
  doi = {10.20382/jocg.v8i2a3},
  copyright = {Copyright (c) 2017 Journal of Computational Geometry},
  langid = {english}
}

@inproceedings{friedlerComparativeStudyFairnessenhancing2019,
  title = {A Comparative Study of Fairness-Enhancing Interventions in Machine Learning},
  booktitle = {Proceedings of the {{Conference}} on {{Fairness}}, {{Accountability}}, and {{Transparency}} - {{FAT}}* '19},
  author = {Friedler, Sorelle A. and Scheidegger, Carlos and Venkatasubramanian, Suresh and Choudhary, Sonam and Hamilton, Evan P. and Roth, Derek},
  year = {2019},
  series = {{{FAT}}* '19},
  eprint = {1802.04422},
  eprinttype = {arxiv},
  address = {{Atlanta, GA, USA}},
  doi = {10.1145/3287560.3287589},
  archiveprefix = {arXiv}
}

@inproceedings{gabillonBestArmIdentification2012,
  title = {Best {{Arm Identification}}: {{A Unified Approach}} to {{Fixed Budget}} and {{Fixed Confidence}}},
  shorttitle = {Best {{Arm Identification}}},
  booktitle = {Advances in {{Neural Information Processing Systems}}},
  author = {Gabillon, Victor and Ghavamzadeh, Mohammad and Lazaric, Alessandro},
  year = {2012},
  volume = {25},
  publisher = {{Curran Associates, Inc.}}
}

@inproceedings{garivierOptimalBestArm2016,
  title = {Optimal {{Best Arm Identification}} with {{Fixed Confidence}}},
  booktitle = {Conference on {{Learning Theory}}},
  author = {Garivier, Aurelien and Kaufmann, Emilie},
  year = {2016},
  pages = {998--1027},
  publisher = {{PMLR}},
  langid = {english}
}

@inproceedings{hashimotoFairnessDemographicsRepeated2018b,
  title = {Fairness {{Without Demographics}} in {{Repeated Loss Minimization}}},
  booktitle = {International {{Conference}} on {{Machine Learning}}},
  author = {Hashimoto, Tatsunori B. and Srivastava, Megha and Namkoong, Hongseok and Liang, Percy},
  year = {2018},
  month = jul,
  eprint = {1806.08010},
  eprinttype = {arxiv},
  publisher = {{PMLR}},
  archiveprefix = {arXiv}
}

@inproceedings{holsteinImprovingFairnessMachine2019a,
  title = {Improving {{Fairness}} in {{Machine Learning Systems}}: {{What Do Industry Practitioners Need}}?},
  shorttitle = {Improving {{Fairness}} in {{Machine Learning Systems}}},
  booktitle = {Proceedings of the 2019 {{CHI Conference}} on {{Human Factors}} in {{Computing Systems}}},
  author = {Holstein, Kenneth and Wortman Vaughan, Jennifer and Daum{\'e}, Hal and Dudik, Miro and Wallach, Hanna},
  year = {2019},
  month = may,
  pages = {1--16},
  publisher = {{ACM}},
  address = {{Glasgow Scotland Uk}},
  doi = {10.1145/3290605.3300830},
  isbn = {978-1-4503-5970-2},
  langid = {english}
}

@inproceedings{jamiesonLilUCBOptimal2014,
  title = {Lil' {{UCB}} : {{An Optimal Exploration Algorithm}} for {{Multi-Armed Bandits}}},
  booktitle = {Conference on {{Learning Theory}}},
  author = {Jamieson, Kevin and Malloy, Matthew and Nowak, Robert and Bubeck, Sebastien},
  year = {2014},
  pages = {17},
  publisher = {{PMLR}},
  langid = {english}
}

@misc{juliaangwinMachineBias2016a,
  type = {Text/Html},
  title = {Machine {{Bias}}},
  author = {Julia Angwin, Jeff Larson},
  year = {2016},
  month = may,
  journal = {ProPublica},
  copyright = {Copyright \textcopyright 2019 ProPublica.},
  howpublished = {https://www.propublica.org/article/machine-bias-risk-assessments-in-criminal-sentencing},
  langid = {english}
}

@inproceedings{kalyanakrishnanPACSubsetSelection2012,
  title = {{{PAC Subset Selection}} in {{Stochastic Multi-armed Bandits}}},
  booktitle = {{{ICML}}},
  author = {Kalyanakrishnan, Shivaram and Tewari, Ambuj and Auer, Peter and Stone, Peter},
  year = {2012},
  pages = {655--662},
  langid = {english}
}

@inproceedings{kaufmannInformationComplexityBandit2013,
  title = {Information {{Complexity}} in {{Bandit Subset Selection}}},
  booktitle = {Conference on {{Learning Theory}}},
  author = {Kaufmann, Emilie and Kalyanakrishnan, Shivaram},
  year = {2013},
  pages = {24},
  publisher = {{PMLR}},
  langid = {english}
}

@inproceedings{kleinbergInherentTradeOffsFair2017,
  title = {Inherent {{Trade-Offs}} in the {{Fair Determination}} of {{Risk Scores}}},
  booktitle = {8th {{Innovations}} in {{Theoretical Computer Science Conference}}},
  author = {Kleinberg, Jon and Mullainathan, Sendhil and Raghavan, Manish},
  year = {2017},
  eprint = {1609.05807},
  eprinttype = {arxiv},
  archiveprefix = {arXiv},
  isbn = {978-3-95977-029-3}
}

@book{lattimoreBanditAlgorithms2020b,
  title = {Bandit {{Algorithms}}},
  author = {Lattimore, Tor and Szepesv{\'a}ri, Csaba},
  year = {2020},
  month = jul,
  edition = {First},
  publisher = {{Cambridge University Press}},
  doi = {10.1017/9781108571401},
  isbn = {978-1-108-48682-8},
  langid = {english}
}

@article{mannorSampleComplexityExploration2004,
  title = {The {{Sample Complexity}} of {{Exploration}} in the {{Multi-Armed Bandit Problem}}},
  author = {Mannor, Shie and Tsitsiklis, John N},
  year = {2004},
  journal = {Journal of Machine Learning Research},
  volume = {5},
  pages = {623--648},
  langid = {english}
}

@article{nargesianTailoringDataSource2021,
  title = {Tailoring Data Source Distributions for Fairness-Aware Data Integration},
  author = {Nargesian, Fatemeh and Asudeh, Abolfazl and Jagadish, H. V.},
  year = {2021},
  month = jul,
  journal = {Proceedings of the VLDB Endowment},
  volume = {14},
  number = {11},
  pages = {2519--2532},
  issn = {2150-8097},
  doi = {10.14778/3476249.3476299},
  langid = {english}
}

@inproceedings{pleissFairnessCalibration2017a,
  title = {On {{Fairness}} and {{Calibration}}},
  booktitle = {Advances in {{Neural Information Processing Systems}}},
  author = {Pleiss, Geoff and Raghavan, Manish and Wu, Felix and Kleinberg, Jon and Weinberger, Kilian Q},
  year = {2017},
  volume = {30},
  publisher = {{Curran Associates, Inc.}}
}

@inproceedings{rolfRepresentationMattersAssessing2021a,
  title = {Representation {{Matters}}: {{Assessing}} the {{Importance}} of {{Subgroup Allocations}} in {{Training Data}}},
  shorttitle = {Representation {{Matters}}},
  booktitle = {{{arXiv}}:2103.03399 [Cs, Stat]},
  author = {Rolf, Esther and Worledge, Theodora and Recht, Benjamin and Jordan, Michael I.},
  year = {2021},
  month = mar,
  eprint = {2103.03399},
  eprinttype = {arxiv},
  primaryclass = {cs, stat},
  archiveprefix = {arXiv},
  langid = {english}
}

@inproceedings{sharmaDataAugmentationDiscrimination2020,
  title = {Data {{Augmentation}} for {{Discrimination Prevention}} and {{Bias Disambiguation}}},
  booktitle = {Proceedings of the {{AAAI}}/{{ACM Conference}} on {{AI}}, {{Ethics}}, and {{Society}}},
  author = {Sharma, Shubham and Zhang, Yunfeng and R{\'i}os Aliaga, Jes{\'u}s M. and Bouneffouf, Djallel and Muthusamy, Vinod and Varshney, Kush R.},
  year = {2020},
  month = feb,
  pages = {358--364},
  publisher = {{ACM}},
  address = {{New York NY USA}},
  doi = {10.1145/3375627.3375865},
  isbn = {978-1-4503-7110-0},
  langid = {english}
}

@article{sheikhiSeparabilityImprecisePoints2017,
  title = {Separability of Imprecise Points},
  author = {Sheikhi, Farnaz and Mohades, Ali and {de Berg}, Mark and Mehrabi, Ali D.},
  year = {2017},
  month = feb,
  journal = {Computational Geometry},
  volume = {61},
  pages = {24--37},
  issn = {09257721},
  doi = {10.1016/j.comgeo.2016.10.001},
  langid = {english}
}

@inproceedings{shekharAdaptiveSamplingMinimax2021,
  title = {Adaptive {{Sampling}} for {{Minimax Fair Classification}}},
  booktitle = {{{arXiv}}:2103.00755 [Cs]},
  author = {Shekhar, Shubhanshu and Fields, Greg and Ghavamzadeh, Mohammad and Javidi, Tara},
  year = {2021},
  month = jul,
  eprint = {2103.00755},
  eprinttype = {arxiv},
  primaryclass = {cs},
  archiveprefix = {arXiv}
}

@inproceedings{taeSliceTunerSelective2021,
  title = {Slice {{Tuner}}: {{A Selective Data Acquisition Framework}} for {{Accurate}} and {{Fair Machine Learning Models}}},
  shorttitle = {Slice {{Tuner}}},
  booktitle = {Proceedings of the 2021 {{International Conference}} on {{Management}} of {{Data}}},
  author = {Tae, Ki Hyun and Whang, Steven Euijong},
  year = {2021},
  month = jun,
  pages = {1771--1783},
  publisher = {{ACM}},
  address = {{Virtual Event China}},
  doi = {10.1145/3448016.3452792},
  isbn = {978-1-4503-8343-1},
  langid = {english}
}

@article{yanProbabilisticConvexHull2015,
  title = {Probabilistic {{Convex Hull Queries}} over {{Uncertain Data}}},
  author = {Yan, Da and Zhao, Zhou and Ng, Wilfred and Liu, Steven},
  year = {2015},
  month = mar,
  journal = {IEEE Transactions on Knowledge and Data Engineering},
  volume = {27},
  number = {3},
  pages = {852--865},
  issn = {1558-2191},
  doi = {10.1109/TKDE.2014.2340408}
}

%%%%%%%%%%%%%%%%%%%%%%%%%%%%%%%%%%%%%%%%%%
\newpage
\begin{appendices}

\section{Proofs}

\subsection{Lower Bounds}
\begin{proof}[Proof of \cref{thm:Oracle Feasible case}]
\label{pf:Oracle Feasible case}
Let $\nu$ be the feasible instance. With the optimal set known, to check feasibility we only need to check the relationship between the mean of each action mean and the set $x_{\epsilon}$. If the set $\mathcal{J}^*$ consists of only one point, it must be sampled enough to determine it lies within $x_{\epsilon}$. If $\mathcal{J}^*$ = \{1,k\}, we must sample to determine that one of the means lies above $x-\epsilon$ and one lies below $x+\epsilon$. 

We start with the case where $\mathcal{J}^* = \{i^*\} \in \{1,k\}$. Since this is a feasible set, it must be that $|p_i - x| < \epsilon$. The closest infeasible case is the boundary of $x_{\epsilon}$ closest to $p_i$. The KL divergence from this infeasible case is given by $\min \left(D(p_i| x-\epsilon), D(p_i | x+\epsilon) \right)$  
    
Let $\tilde{t}_i = \max \left\{ D(p_i|x-\epsilon)^{-1}, D(p_i|x+\epsilon)^{-1} \right\} \frac{1}{2} \log(\frac{1}{4\delta})$. We let $N_i$ be the random variable representing the number of times action $i$ was sampled when the policy terminates. We will use a proof by contradiction similar to that presented by \cite{mannorSampleComplexityExploration2004} along with a divergence decomposition \cite[Lemma 15.1]{lattimoreBanditAlgorithms2020b}.
    
Assume $E[N_{i^*}] \leq \tilde{t}_{i^*}$. 
Let $O \in \{feasible, infeasible\}$ be the output of a policy, and define event $B = \{O = feasible\}$. Then by definition of a $1-\delta$-sound policy, $P_{\nu}(B) \geq 1-\delta \geq 1/2$ for $\nu \in \mathcal{E}_{f}$. Without loss of generality, assume $x - \epsilon<p_{i^*}\leq x$. Then the closest infeasible case would be $p_{i^*} = x-\epsilon$. We will call $H_0: \ p_{i^*} = p_{i^*}, \ H_1: p_{i^*} = x-\epsilon$. We get that,

\begin{align*}
    P_{1}(B) &= E_{1}[1\{B\}]\\
    &= E_{0}\left[\frac{L_{1}}{L_0}1\{ B\}\right]\\
    &= E_{0}\left[\frac{L_{1}}{L_0} | B \right] P_0(B)\\
    &=E_{0}\left[\exp\left\{-\log \left(\frac{L_{0}}{L_1} \right)\right\} | B\right]P_0(B)\\
    &\geq \exp \left\{- E_{0}\left[ \log \left(\frac{L_{0}}{L_1} \right) | B\right] \right\}P_0(B)\\
     &= \exp \left\{- E_{0}\left[ N_i | B\right] D(p_{i^*}, x-\epsilon) \right\}P_0(B)\\
     &\geq \exp \left\{- 2\tilde{t}_{i^*} D(p_{i^*}, x-\epsilon) \right\}P_0(B)\\
     &= 4\delta P_0(B)\\
     &> \delta
\end{align*}

which contradicts that the policy is $1-\delta$ sound under hypothesis $H_{1}$, which means that 

\begin{equation*}
    E_0[N_{i^*}] \geq \max \left\{ D(p_{^*}|x-\epsilon)^{-1}, D(p_{i^*}|x+\epsilon)^{-1} \right\} \frac{1}{2} \log\left(\frac{1}{4\delta}\right).
\end{equation*}

For $\mathcal{J}^* = \{1,k\}$, we define $H_0: \ p_1 = p_1, p_k = p_k$ and $H_1:\ P_1 = x - \epsilon \text{ or } p_k = p+\epsilon$. Because $p_1\geq p_k$ by definition, if either $p_1 = x-\epsilon$ or $p_k=x+\epsilon$ then the problem is infeasible. By setting $\tilde{t}_i = \max \left\{ D(p_i|x-\epsilon)^{-1}, D(p_i|x+\epsilon)^{-1} \right\} \frac{1}{2} \log(\frac{1}{4\delta})$ and following the same method as above for actions $i \in \{1, k\}$, get 

\[
E_0[T_i] \geq \min \left\{ D(p_i|x-\epsilon)^{-1}, D(p_i|x+\epsilon)^{-1} \right\} \frac{1}{2} \log\left(\frac{1}{4\delta}\right).
\]
\end{proof}

\begin{proof}[Proof sketch of
\cref{thm:Infeasible case}]
\label{pf:Infeasible case}
The proof for the infeasible lower bound follows closely to that of the feasible case, therefore we provide a brief proof outline. Because all means must lie outside $x_{\epsilon}$, the closest feasible  case is the boundary of $x_{\epsilon}$ nearest the  means. To determine infeasibility, all actions must be sampled sufficiently to reject this boundary. Without loss of generality, if we assume $p_i < x - \epsilon$, then the closest boundary would be $x - \epsilon$. Setting $H_i: p_i = p_i$, $H_1: p_i = x - \epsilon$ for all $i$, $\tilde{t}_i = \max \left\{ D(p_i|x-\epsilon)^{-1}, D(p_i|x+\epsilon)^{-1} \right\} \frac{1}{2} \log(\frac{1}{4\delta})$ and following the methods from the feasible case, we get the desired result.
\end{proof}

\subsection{Upper Bounds}

\begin{proof}[Proof of \cref{thm:uniform}]
\label{pf:uniform}
Given some $\delta$ and $B(n, \delta)$ that satisfies \cref{eq:margin}, let event $E$ be the event that all confidences regions contain their mean, $E = \{\forall i \in [k], n \in \mathbb{N}, \ \hat{p}_i(n) - B(n, \delta) \leq p_i \leq \hat{p}_i(n) + B(n, \delta)\}$. Under event $E$, each action $i$ will become inactive at or before being sampled $s_i^{min}$ times. 

We start with the feasible cases. When where $\mathcal{J}^* = \{i^*\}$ and under event $E$, action $i^*$ can be sampled at most $s_{i^*}^{min}$ times before the policy will terminate due to stopping rule \ref{sr:fB}. Thus the bound on the sample size of each action is the minimum of the sample size it is guaranteed to become inactive under $E$, which is $s_i^{min}$, and the sample size of $s_{i^*}^{min}$ when the policy terminates. Since event $E$ happens with probability at least $1-\delta$, this concludes the proof when the optimal subset is one action.

In the case where $\mathcal{J}^* = \{1,k\}$, under event $E$ the policy will terminate due to stopping rule \ref{sr:fB}, which will happen when action 1 is sampled $s_1^{max}$ times and action $k$ is sampled $s_k^{max}$ times. Again, under $E$ an action becomes inactive when sampled at most $s_i^{min}$ times, this gives that each action is sampled at most  $\left( s_i^{min}, \max \left(s_1^{max}, s_k^{max} \right) \right)$ times. Again, event $E$ happens with probability at least $1-\delta$.

When the problem is infeasible, each action will be sampled until its confidence region is disjoint form $x_{\epsilon}$. Under event $E$, this sample size is bounded above by $s_i^{min}$ for all $i$. 
\end{proof}

\begin{proof}[Proof of \cref{thm:LUCBf}]
\label{pf:LUCBf}
We Start with the feasible case where there exists a mean on both sides of $x$. Let event $E$ be the event that all confidences regions contain their mean, $E = \{\forall i \in [k], n \in \mathbb{N}, \ \hat{p}_i(n) - B(n, \delta) \leq p_i \leq \hat{p}_i(n) + B(n, \delta)\}$. 
Without loss of generality, let actions $i,j$ be the action that triggers termination at time $\tau$. Define the sample size of action $l$ at time $t$ as $N_l(t)$. Assume without loss of generality that $j^* = k$, and $p_j < x$, thus $i^* = 1$ and $p_i > x$. If $j=k$, then under event $E$, $N_j(\tau) \leq s_k^{max}$. If $j\neq k$ it must be that, 
\begin{align*}
    & \hat{p}_j(N_j(\tau)-1) - B(N_j(\tau)-1) \leq p_k & \text{by $E$}\\
    & \hat{p}_j(N_j(\tau)-1) + B(N_j(\tau)-1) \geq x + \epsilon &\text{definition of } \tau
\end{align*}
Therefore 

\begin{align*}
    \hat{p}_j(N_j(\tau)-1) + B(N_j(\tau)-1) &\geq x + \epsilon\\
    2B(N_j(\tau)-1) &\geq x + \epsilon - (\hat{p}_j(N_j(\tau)-1) - B(N_j(\tau)-1))\\
    &\geq x + \epsilon- p_k\\
    & = \Delta_k^{max}\\
    &> 2B(s_k^{max})
\end{align*}
since $B$ is a decreasing function, $N_j(\tau) - 1 < s_k^{max} \implies N_j(\tau) \leq s_k^{max}$. Similarly for action $i$ we have that $ N_j(\tau) \leq s_1^{max}$

For non-terminating actions we have that under $E$,
\begin{align*}
   &\hat{p}_l(N_l(\tau)-1) - B(N_l(\tau)-1) \leq p_k 
   &  \hat{p}_l(N_l(\tau)-1) + B(N_l(\tau)-1) \geq \max(x+\epsilon, p_l)
\end{align*}
which implies that 
\[
2B(N_l(\tau)-1)\geq \max (|p_k - (x+\epsilon)|, |p_l - p_k|) = \max (\Delta_k^{max}, \Delta_{l,j}) > \max(2B(s_k^{max}), 2B(s_{k,l}))
\]
giving $N_l(\tau) \leq \min(s_k^{max}, s_{k,l})$ and similarly $N_l(\tau) \leq \min(s_1^{max}, s_{1,l})$. To meet both these bounds, it must be that $N_l(\tau) \leq \max ( \min(s_1^{max}, s_{1,l}), \min(s_k^{max}, s_{k,l}) ) = \max ( \min(s_{i^*}^{max}, s_{l,i^*}), \min(s_{j^*}^{max}, s_{l,j^*}) )$. 

When $s_{j^*}^{max} \leq s_{l,j^*}$, which is equivalent to $\Delta_{j^*}^{max} \geq \Delta_{l,j^*} $, then $ s_{j^*}^{max} \geq s_{i^*}^{max}$ by definition. If $s_{i^*}^{max} \geq s_{l,i^*}$, then we have that 
   
\begin{align*}
    \Delta_{l,i^*} &= \Delta_{i^*} + \Delta_l^{min}\\
    &\geq \Delta_{j^*}^{max} + \Delta_l^{min}\\
    &\geq \Delta_{j^*}^{max}
\end{align*}
Thus $s_{j^*}^{max} \geq s_{l,i^*}$.
So when $\Delta_{j^*}^{max} \geq \Delta_{l,j^*}$, the sample size of action $l$ is bounded above by $s_{j^*}^{max}$.

When $s_{j^*}^{max} > s_{l,j^*}$, which is equivalent to $\Delta_{j^*}^{max} < \Delta_{l,j^*}$, it must be that $p_{i^*}$ and $p_l$ are on the same side of $x_{\epsilon}$, thus $s_{i^*}^{max} < s_{l,i^*}$ and the sample size of action $i$ is bounded above by $\max(s_{l,j^*}, s_{i^*}^{max})$.

When $\mathcal{J}^* = {l^*}$. there are two scenarios. Either all means lies in $x_\epsilon$, or all means lie in one direction from $x_\epsilon$. In the first case, the outcome is the same as above. In the second case, it must be that the optimal action mean must be sample at most $s_{l^*}^{min}$ times, at which point it would trigger termination under $E$. Let $j$ be the action that triggered stopping rule \ref{sr:fB}. Without loss of generality, assume $p_{l^*} <x$. If $j=l^*$, then under $E$, $N_j(\tau) \leq s_{l^*}^{min}$. If $j\neq l^*$, 
\begin{align*}
    & \hat{p}_j(N_j(\tau)-1) + B(N_j(\tau)-1) \geq p_{l^*} & \text{by $E$}\\
    & \hat{p}_j(N_j(\tau)-1) - B(N_j(\tau)-1) \leq x - \epsilon &\text{definition of } \tau
\end{align*}
and as in the feasible case,

\begin{align*}
    \hat{p}_j(N_j(\tau)-1) - B(N_j(\tau)-1) &\leq x - \epsilon\\
    2B(N_j(\tau)-1) &\geq \hat{p}_j(N_j(\tau)-1) - B(N_j(\tau)-1)  - (x + \epsilon)\\
    &\geq p_{l^*} - (x - \epsilon)\\
    & = \Delta_{l^*}^{min}\\
    &> 2B(s_{l^*}^{min})
\end{align*}
which gives that $N_j(\tau) \leq s_{l^*}^{min}$. 
For all other actions $l\neq j$,
\begin{align*}
  &\hat{p}_l(N_l(\tau)-1) + B(N_l(\tau)-1) \geq p_j 
  &  \hat{p}_l(N_l(\tau)-1) - B(N_l(\tau)-1) \leq \min(x-\epsilon, p_l)
\end{align*}
and using the same logic from above gives $N_l(\tau) \leq \min (s_{l^*}^{min}, s_{l,l^*})$. 

In the infeasible case, assume without loss of generality that $p_i \leq x_{\epsilon}$ for all $i \in [k]$. Under $E$, it must be that $\hat{p}_i(N_i(\tau)-1) - B(N_i(\tau)-1) \leq p_i$ and $\hat{p}_i(N_i(\tau)-1) - B(N_i(\tau)-1) \geq x - \epsilon$ for all $i \in [k]$. Therefore 
    
\begin{align*}
    2B(N_i(\tau)-1) &\geq x-\epsilon - p_i\\
    & = \Delta_i^{min}\\
    &> 2B(s_i^{min})
\end{align*}
and $N_i(\tau) \leq s_i^{min}$. Since $E$ happens with probability at least $1-\delta$, this concludes the proof. 
\end{proof}

\subsection{Upper bound improvement of LUCB Mean over Uniform}
\label{ap:LUCB-U}

We start with the case where $\exists i, j, \ p_i<x<p_j$, $|\mathcal{J}^*|=2$. If $\Delta_{l,j^*} \leq \Delta_{j^*}^{max}$ then $\min(s_{j^*}^{max}, s_l^{min}) \geq s_{j^*}^{max}$, because
\begin{align*}
\Delta_l^{min} &\leq 
\begin{cases}
\Delta_{j^*}^{min}
& p_l \notin x_{\epsilon}
\\
\epsilon
& p_l \in x_{\epsilon}
\end{cases}
\\
 & < \Delta_{j^*}^{max}
\end{align*}
Therefore $\min(s_{j^*}^{max}, s_l^{min}) \geq s_{j^*}^{max}$ when $\Delta_{l,j^*} \leq \Delta_{j^*}^{max}$. 

If $\Delta_{l,j^*} > \Delta_{j^*}^{max}$, then $p_l$ is on the same side of $x_{\epsilon}$ as $p_{i^*}$. To show $\min(s_{j^*}^{max}, s_l^{min}) \geq \max(s_{i^*}^{max}, s_{l,j^*})$, we have that, 

\begin{align*}
    \Delta_{l}^{min} &< \Delta_{j^*}^{max} + \Delta_{l}^{min} = \Delta_{l,j^*}\\
\Delta_{l}^{min} &< \Delta_{i^*}^{max} & \text{by definition}
\end{align*}
and by definition
\begin{align*}
    \Delta_{j^*}^{max} &< \Delta_{i^*}^{max}\\
    \Delta_{j^*}^{max} &< \Delta_{l,j^*}
\end{align*}

Since $\min(s_{j^*}^{min}, s_l^{min}) \geq \min(s_{j^*}^{max}, s_l^{min}) \geq \max(s_{i^*}^{max}, s_{l,j^*})$ this covers the $|\mathcal{J}^*|=1$ case as well. 

When all means are on one side of $x$, then $|\mathcal{J}^*|=1$ and must show $\min(s_{j^*}^{min}, s_l^{min}) \geq \min(s_{l,j^*}, s_{j^*}^{min})$. If  $\Delta_{j^*}^{max} \leq \Delta_l^{min}$ then we have,
\begin{align*}
    \Delta_l^{min} &< \Delta_{j^*}^{min} + \Delta_{l^*}^{min} = \Delta_{l,j^*}
\end{align*}
We have therefore shown in all cases, LUCB Mean has a lower high probability upper bound on sample complexity in the $(x,\epsilon)$-feasible Bernoulli setting than Uniform.

\end{appendices}
\end{document}